\documentclass[10pt,twocolumn,letterpaper]{article}

\usepackage{iccv}
\usepackage{times}
\usepackage{epsfig}
\usepackage[lined, boxed, linesnumbered, commentsnumbered]{algorithm2e}
\usepackage{graphicx}
\usepackage{amsmath}
\usepackage{amssymb,verbatim,array}
\usepackage{color}
\usepackage[]{algorithm2e}

\usepackage[pagebackref=true,breaklinks=true,letterpaper=true,colorlinks,bookmarks=false]{hyperref}

\iccvfinalcopy 

\def\iccvPaperID{2610} 

\ificcvfinal\fi
\begin{document}

\title{Fast Non-local Stereo Matching based on Hierarchical Disparity Prediction}

\author{\textbf{Xuan Luo, Xuejiao Bai, Shuo Li, Hongtao Lu}\\
Shanghai Jiao Tong University,\\
No. 800, Dongchuan Road, Shanghai, China\\
{\tt\small \{roxanneluo, yukiaya, uyas, htlu\}@sjtu.edu.cn}
\and
\textbf{Sei-ichiro Kamata}\\
Waseda University\\
{\tt\small kam@waseda.jp}
}

\maketitle

\begin{abstract}
    Stereo matching is the key step in estimating depth from two or more images. Recently, some tree-based non-local stereo matching methods \cite{MST,ST} have been proposed, which achieved state-of-the-art performance. The algorithms employed some tree-structures to aggregate cost and thus improved the performance and reduced the coputation load of the stereo matching. However, the computational complexity of these  tree-based algorithms is still high because they search over the entire disparity range. In addition, the extreme  greediness of the minimum spanning tree (MST) causes the poor performance in large areas with similar colors but varying disparities. In this paper, we propose an efficient stereo matching method using a hierarchical disparity prediction (HDP) framework to dramatically reduce the disparity search range so as to speed up the tree-based non-local stereo methods. Our disparity prediction scheme works on a graph pyramid derived from an image whose disparity to be estimated. We utilize the disparity of a upper graph to predict a small disparity range for the lower graph. Some independent disparity trees (DT) are generated to form a disparity prediction forest (HDPF) over which the cost aggregation is made. When combined with the state-of-the-art tree-based methods, our scheme not only dramatically speeds up the original methods but also improves their performance by alleviating the second drawback of the tree-based methods. This is partially because our DTs overcome the extreme greediness of the MST. Extensive experimental results on some benchmark datasets demonstrate the effectiveness and efficiency of our framework. For example, the segment-tree based stereo matching becomes about 25.57 times faster and $2.2\%$ more accurate over the Middlebury 2006 full-size dataset \cite{Midd4}.
\end{abstract}

\section{Introduction}

\label{sec:intro}
Stereo matching has been one of the most challenging problems in computer vision. 
It takes as input two images that are taken from different views of a single scene, in a manner similar to human binocular vision, and matches pixels in the two images to obtain their visual disparities. Then, the depth information can be  extracted from the disparity (which is inversely proportional to the depth).

A variety of binocular stereo matching algorithms have been proposed in recent years. 
Generally, the stereo matching methods can be divided into two broad categories: global and local \cite{taxonomy}.

Global algorithms compute the disparity of each pixel by using the disparity estimates of all the other pixels. These methods can typically be formulated as an energy-minimization problem and can be solved with some optimization techniques. The estimated disparity can be obtained as a optimal solution to the energy function. 
Typical optimization approaches include graph cut \cite{GraphCut}, loopy belief propagation \cite{Belief1,Belief2} and dynamic programming \cite{Dynamic1,Dynamic2,Tree}. However, these methods achieve high accuracy at the expense of long runtime and large memory space. Therefore, it is difficult to apply them in practice even with the help of GPU (graphics processing unit) \cite{GPU}.

Local algorithms, on the other hand, are of much higher efficiency. In local methods, the disparity of each	 pixel only relies on a local support window instead of the whole image. The local algorithms are also considered as the cost-aggregation-based methods \cite{Cost0,Cost1,Cost2,Cost3,Cost4}. Generally, they contain four steps: 1) matching cost computation, where the dissimilarity (cost) of corresponding pair of pixels is computed for all possible disparities; 2) cost aggregation, where the matching cost is aggregated over a support window around each pixel to reduce noise; 3) disparity computation, where for each pixel, an optimal disparity with the lowest aggregated cost is selected as the estimated disparity value, the winner-takes-all scheme is often adopted; 4) disparity refinement, which further improves the accuracy of disparities. The local methods are usually less accurate than the global ones.

To improve the accuracy of the local methods, Yoon and Kweon \cite{JBF} proposed a local weighting approach.Instead of using square windows with uniform weights, the weighting technique aggregates matching cost over a window based on color similarity and spatial distance, resembling of a joint bilateral filter. Let $C(\mathbf{x},d)$ denote the cost function of pixel $\mathbf{x}$ at disparity $d$, and $C^A(\mathbf{x},d)$ denote the aggregated cost function. $C^A(\mathbf{x}_i,d)$ of pixel $\mathbf{x}_i$ is usually expressed as a convex combination of $C(\mathbf{x}_j,d)$ over a window centered at $\mathbf{x}_i$,
\begin{eqnarray}
C^A(\mathbf{x}_i,d)=\sum_{j}w_{ij}C(\mathbf{x}_j,d)   \label{aggre}
\end{eqnarray}  
where the weight $w_{ij}$ is given by 
\begin{eqnarray}
w_{ij}=\frac{1}{K_i} \mathrm{exp}(-\frac{||\mathbf{x}_i-\mathbf{x}_j||^2}{\sigma_s^2})\mathrm{exp}(-\frac{||I_i-I_j||^2}{\sigma_r^2})
\end{eqnarray}  
and $\mathbf{x}$ is the pixel coordinate and $I$ is the intensity or color of the image, $\sigma_s$ and $\sigma_r$ are two parameters to control the similarities contributed by spatial and intensity relations, respectively. $K_i$ is a normalized parameter such that $\sum_j w_{ij} =1$. Since then, many bilateral filters \cite{BF1,BF2,FastBF} have been used to further improve the accuracy. 
However, the full-kernel implementation of the bilateral filter is still slow.

To further accelerate the bilateral filter, a number of approximation approaches have been proposed. Paris and Durand proposed a fast bilateral filter \cite{FastBF} implemented on GPU. Porikli developed an O(1) bilateral filter \cite{ConstBF}. However, their accuracy is lower than that of the full-kernel implementation \cite{BFDemo}. He {\it et al.} \cite{GuidedBF} developed the guided image filter for stereo matching. Its runtime is linear in the number of image pixels and was demonstrated to outperform all the other local methods on Middlebury datasets\cite{MiddEval} on both efficiency and accuracy \cite{Cost2}.

Recently, Yang proposed a non-local cost aggregation approach \cite{MST}, which is faster and more accurate than the guided image filter.
This algorithm combines the advantages of both the local and the global methods. The reference image is treated as a 4-connected undirected planar graph where each pixel corresponds to a node and each pair of neighboring pixels is connected by an edge. A minimum spanning tree (MST) is built based on the weight (color similarity between neighboring pixels) of each edge. For each pair of nodes, their shortest distance on the MST decides their similarity.
The method aggregates the matching cost over the constructed MST structure. Specifically, the weight between a pair of neighboring pixels $r$ and $s$ is 
\begin{eqnarray}
w(r,s)=|I(r)-I(s)|
\end{eqnarray}    
When a minimum spanning tree (MST) has been computed from the graph, the similarity between two pixels is defined in terms of the MST by the length of their shortest path in the MST. Let $D(i,j)$ denote the distance of two pixels $\mathbf{x}_i$ and $\mathbf{x}_j$ in the MST, the similarity (weight) $w_{ij}$ between them  is given by 
\begin{eqnarray}
w_{ij}= \mathrm{exp}(-\frac{D(i,j)}{\sigma})
\end{eqnarray} 
where $\sigma$ is a parameter. With these weights, the cost aggregation on the MST is conducted using Eq. (\ref{aggre}).       
As demonstrated in \cite{MST}, the MST-based stereo matching outperfoms all the local methods. 
 
Based on the MST, segment-tree (ST) \cite{ST} further incorporates segmentation technique into the MST framework and has be shown to achieve better performance.

MST and ST are recognized as non-local methods as every pixel can receive supports from all the other pixels in the whole image.
Their support windows can be freely extended by making pixels of similar colors close on the tree.
They overcome the drawbacks of traditional local stereo algorithms in selecting eligible support windows.
Besides, a linear time exact algorithm is proposed to aggregate the matching cost over the tree structure, where only 2 addition/subtraction operations and 3 multiplication operations are required for each pixel at each possible disparity \cite{MST}. This is very close to the complexity of the most efficient unnormalized box filtering using integral image \cite{BoxFilter} so it sharply decreases the complexity of edge-aware filter and the cost aggregation.

However, there are still two disadvantages in these tree-based algorithms.
\begin{enumerate}
   \item The computational complexity of the tree-based algorithms is still high because they search over the entire disparity range.
   \item The extreme greediness of MST causes its poor performance in large areas with similar colors but varying disparities since pixels in these areas are joined too close to discern their disparity differences.
\end{enumerate}

The drawbacks of such high computation cost and greediness of the tree-based algorithms motivate us to propose new strategies to reduce the disparity search range and improve the accuracy as well. 
The main idea of our method is to construct a hierarchical disparity prediction framework, based on image pyramid, that can predict a small disparity interval within which the true disparity falls with high probability so that the high computation cost can be largely reduced. Moreover, the predicted disparity intervals help to better segment the image. So unlike typical stereo matching methods which use color similarity to approximate disparity similarity, our segmentation uses disparity similarity directly, and thus gives rise to higher accuracy.

\section{Hierarchical Disparity Prediction}
Although the tree-base non-local stereo matching methods MST and ST achieved better performance than the traditional methods, they still suffer from some drawbacks as stated in Section 1. Particularly the computation efficiency is a bottleneck for stereo matching in real applications, especially when the image is of big size. The time complexity of the tree-based stereo marching methods is $O(nd)$ where $n$ is the number of pixels in the image and $d$ is the maximum possible disparity value. The focus of this paper is to reduce the complexity induced by large $d$ by predicting disparity in a hierarchical way. 

\subsection{Some Key Observations}
Our hierarchical disparity prediction (HDP) is based on a hierarchical graph pyramid structure.   

\begin{figure}[t]
	\begin{minipage}[t]{0.49\linewidth}
		\centerline{\includegraphics[ width = \linewidth]{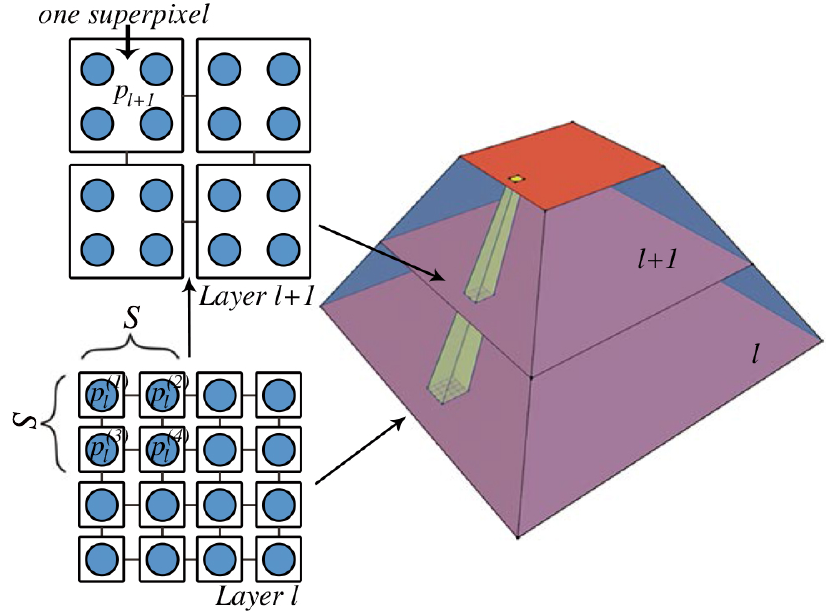}}
		\caption {Graph Pyramid construction between $G_l$ and $G_{l+1}$}
		\label{fig:layer_pri}
	\end{minipage}
	\begin{minipage}[t]{0.49\linewidth}
		\centerline{\includegraphics[ width = \linewidth]{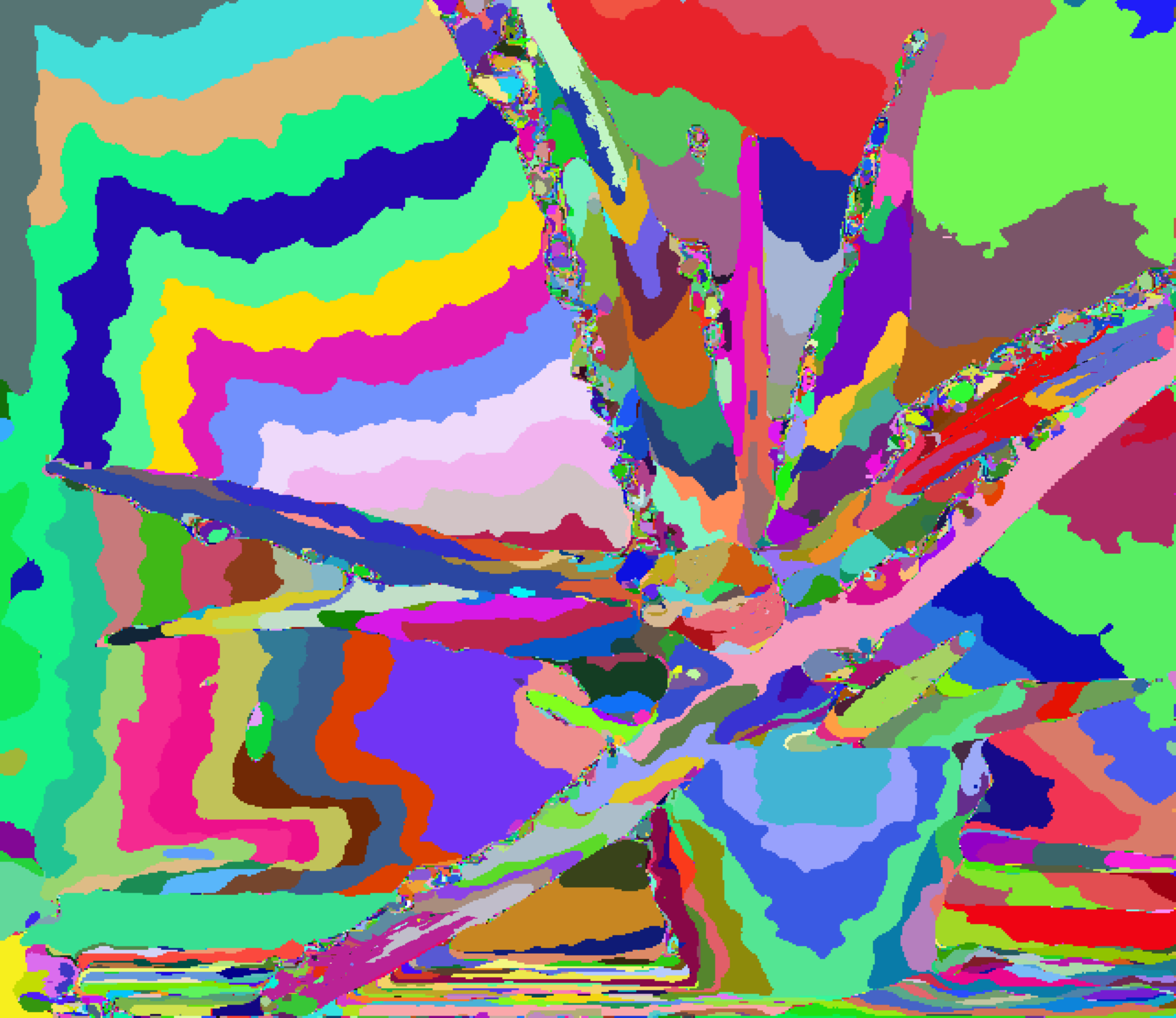}}
		\caption{HDPF of HDP+MST over full-size Aloe in \cite{Midd4}.}
		\label{fig:forest-segment}
	\end{minipage}
\end{figure}

\begin{figure}[t]
	\begin{minipage}[t]{0.52\linewidth}
		\centerline{a.\includegraphics[ width = 3.9cm]{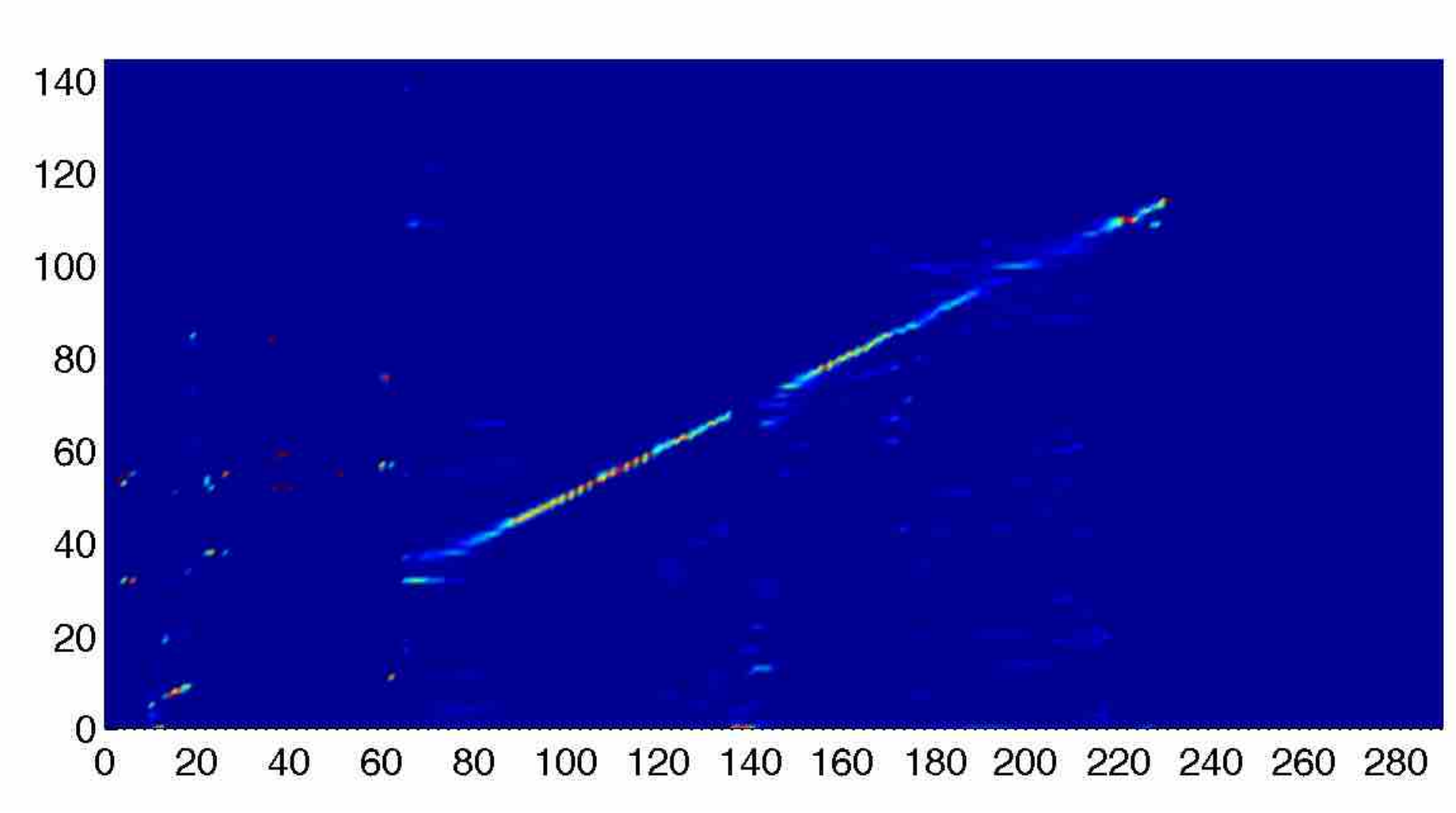}}
	\end{minipage}
	\begin{minipage}[t]{0.46\linewidth}
		\centerline{\includegraphics[ width = 3.9cm]{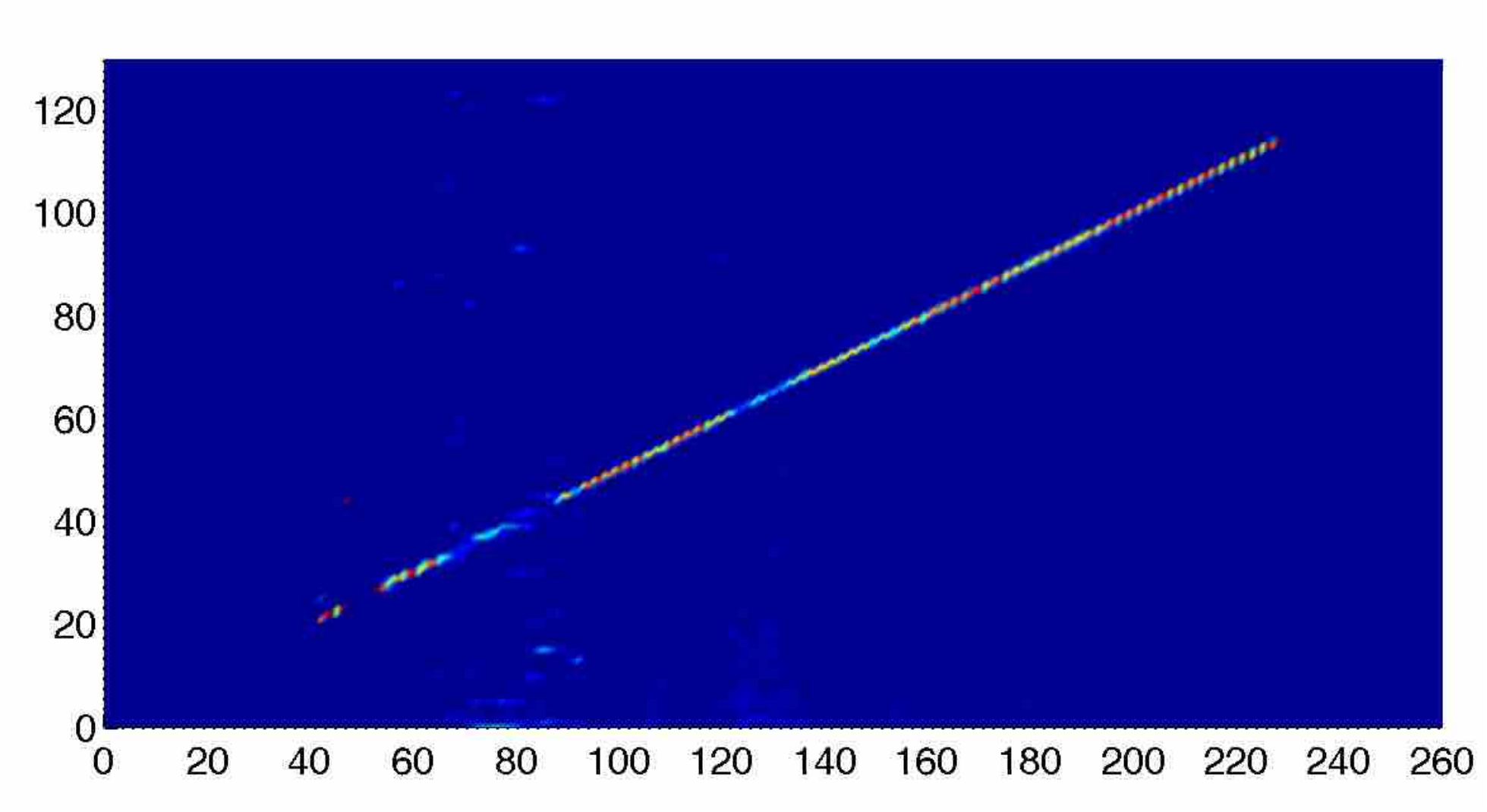}}
	\end{minipage}
	
	\begin{minipage}[t]{0.52\linewidth}
		\centerline{b.\includegraphics[ width = 3.9cm]{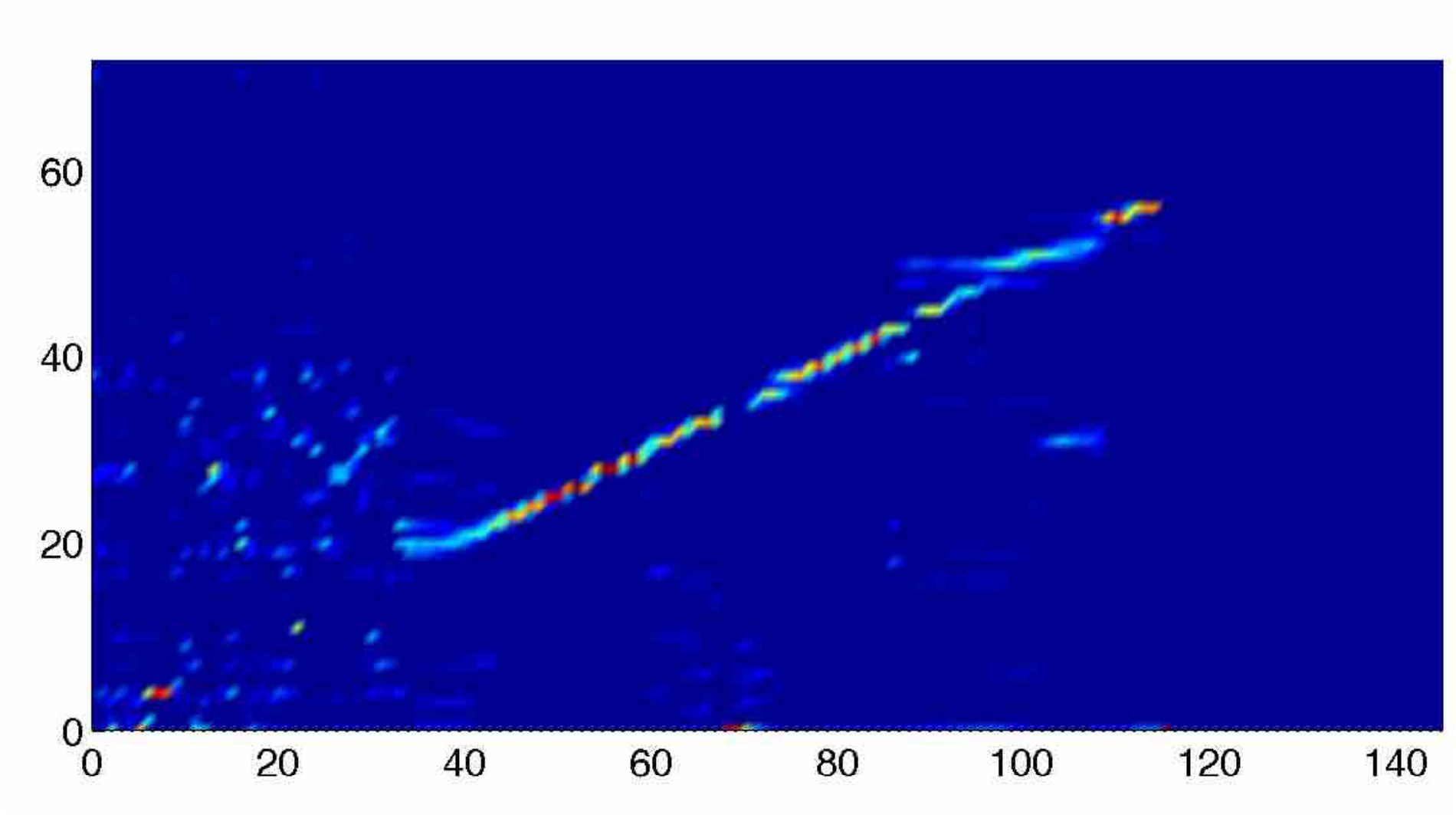}}
	\end{minipage}
	\begin{minipage}[t]{0.46\linewidth}
		\centerline{\includegraphics[ width = 3.9cm]{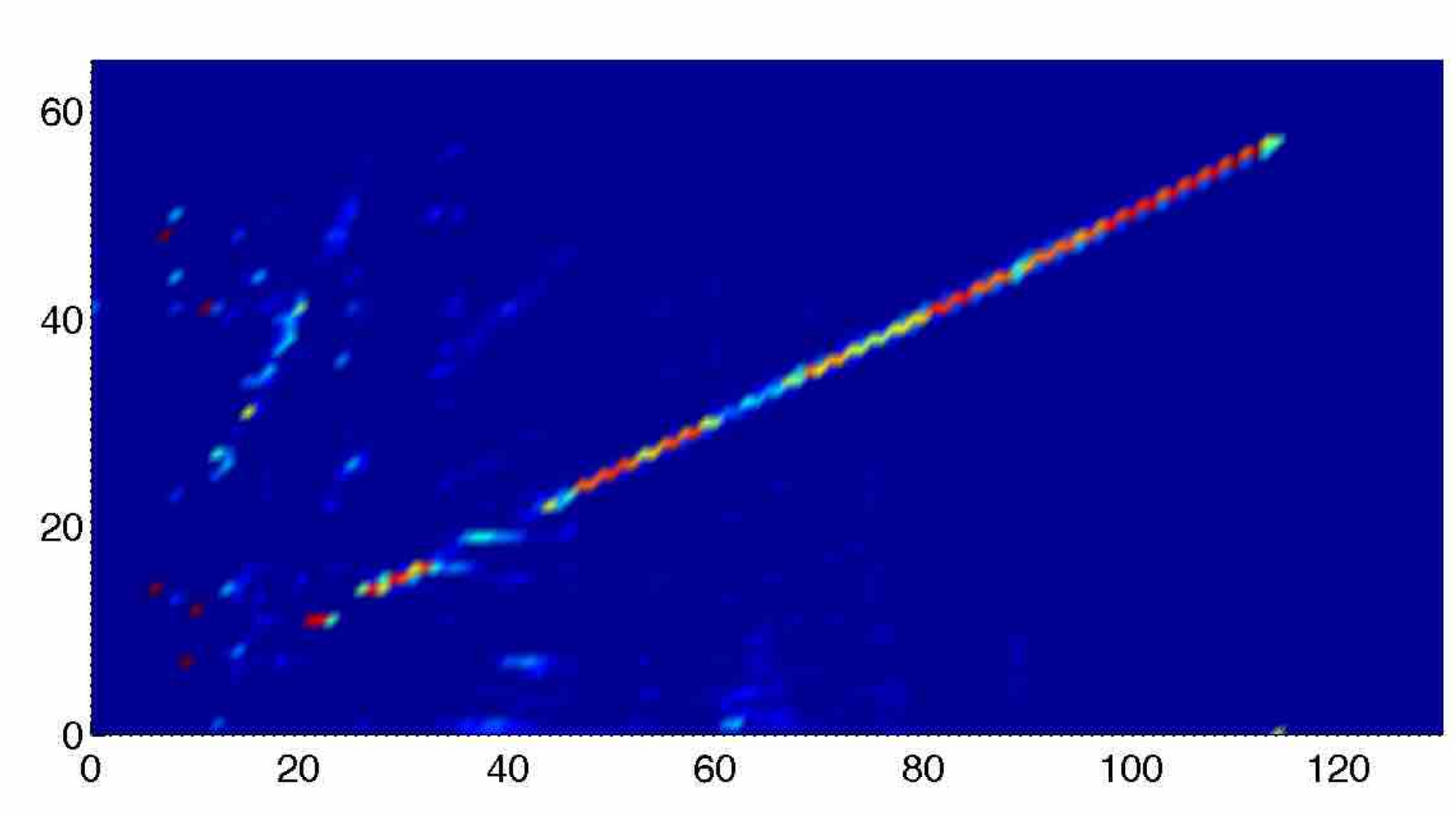}}
	\end{minipage}
	
	\begin{minipage}[t]{0.52\linewidth}
		\centerline{Bowling1}
	\end{minipage}
	\begin{minipage}[t]{0.46\linewidth}
	\centerline{Colth2}
    \end{minipage}
\caption{True distributions $\mathbf{P}(\mathbf{D}_{l+1}|\mathbf{D}_l)$ for a. $l=0$  and b. $l=1$ for Bowling1 and Cloth2 in Middlebury 2006 dataset with full-size resolution.}
\label{fig:sgl}
\end{figure}

\begin{figure}
	
	\begin{minipage}[b]{0.52\linewidth}
		\centerline{a.\includegraphics[ width = 3.9cm]{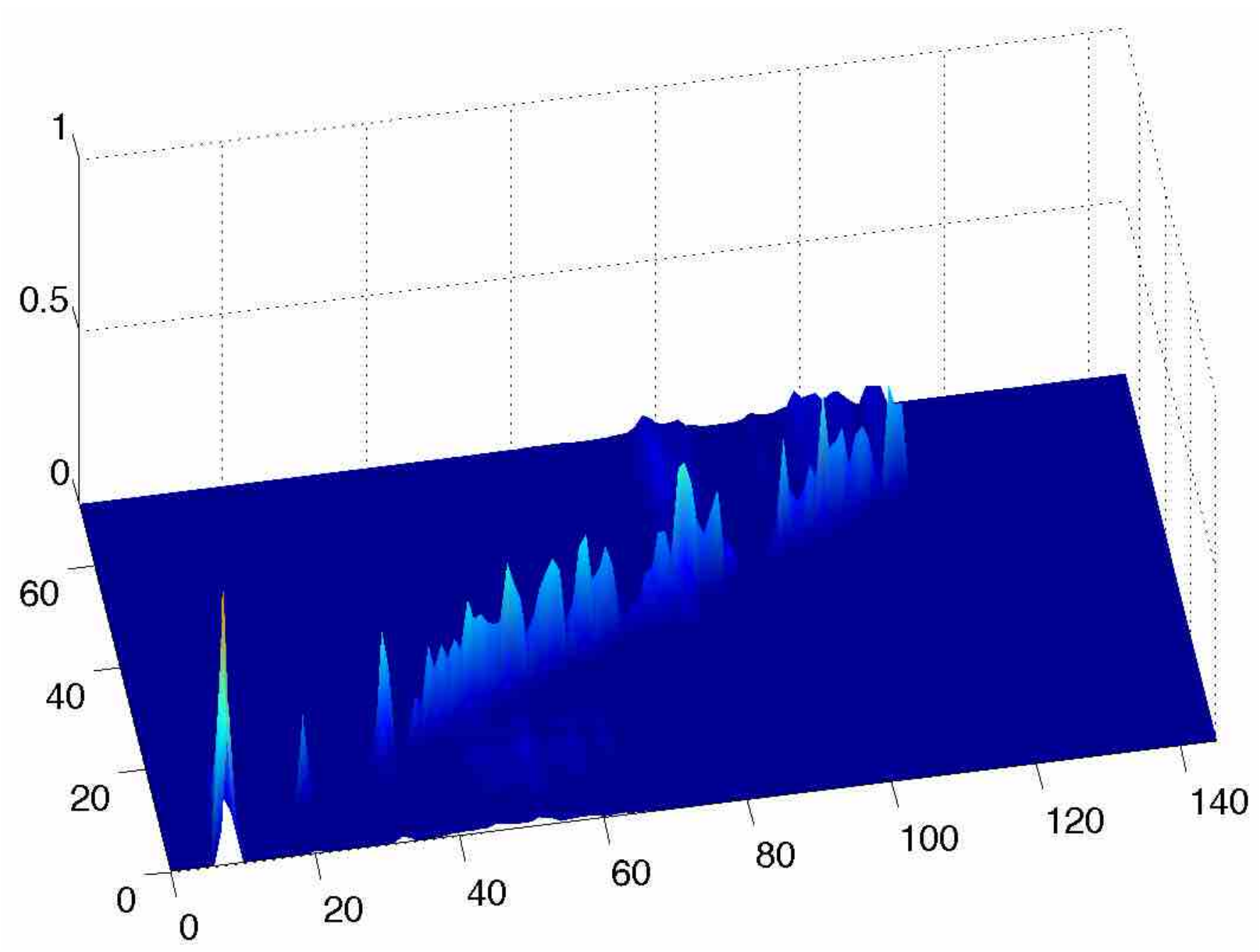}}
	\end{minipage}
	\begin{minipage}[b]{0.46\linewidth}
		\centerline{\includegraphics[ width = 3.9cm]{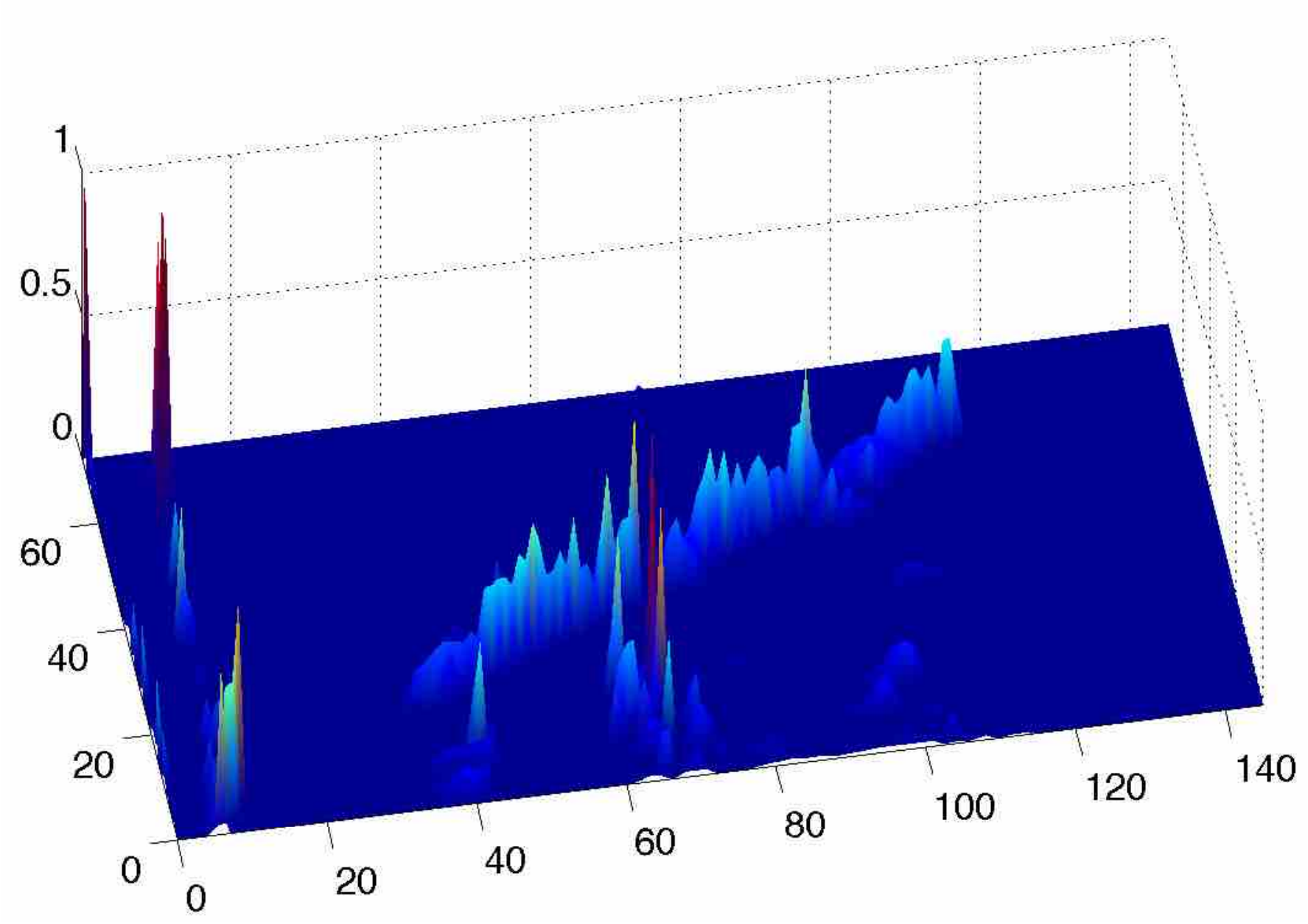}}
	\end{minipage}
	\begin{minipage}[b]{0.52\linewidth}
		\centerline{b.\includegraphics[ width = 3.9cm]{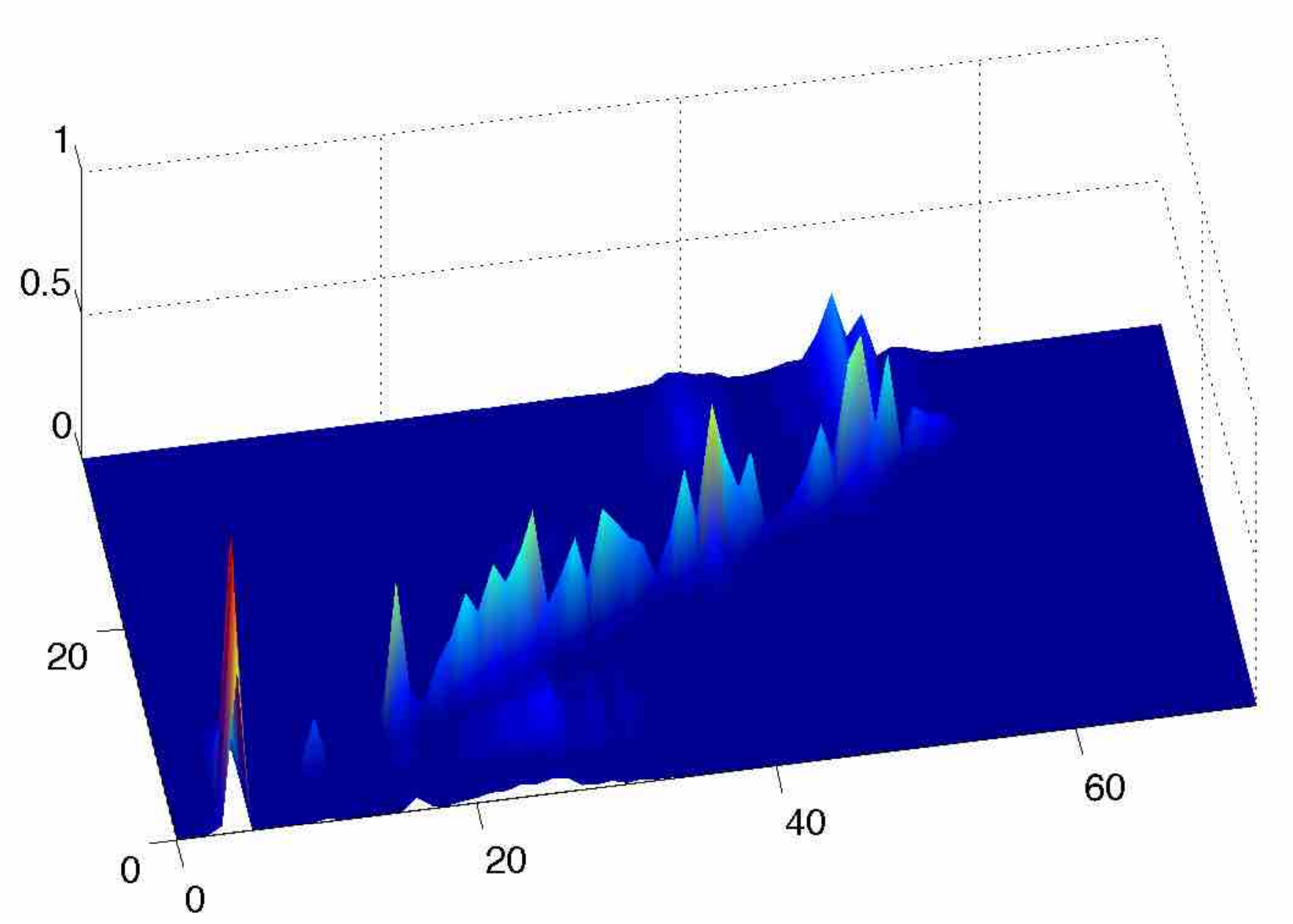}}
	\end{minipage}
	\begin{minipage}[b]{0.46\linewidth}
		\centerline{\includegraphics[ width = 3.9cm]{pic/picked-matrix-pic/halfsize/lgs/Bowling1-true-statistical-Pr_D0-D1.pdf}}
	\end{minipage}

	\begin{minipage}[t]{0.46\linewidth}
		\centerline{\small Predicted }
	\end{minipage}
	\begin{minipage}[t]{0.46\linewidth}
		\centerline{\small Sampled }
	\end{minipage}
	\caption{Comparisons between the sampled and our predicted $\mathbf{P}(\mathbf{D}_l|\mathbf{D}_{l+1})$ of a. $l=0$, b. $l=1$ for Bowling1 in Middlebury 2006 dataset with half-size resolution.}
	\label{fig:lgs}
\end{figure}

First, we build two graph pyramids each for the left and right input image. A  pyramid consists of $L+1$ layers, two layers of it are shown in Fig.~\ref{fig:layer_pri}. Each layer is represented by a standard 4-connected graph $G_l (V_l , E_l )$ where $V_l$ is the set of nodes and $E_l$ is the set of edges connecting neighboring nodes, $l=0,\dots,L$. The hierarchical graphs are constructed recursively from the lowest layer $G_0$, which is the image itself. Fig.~\ref{fig:layer_pri} illustrates how $G_{l+1}$ is constructed from $G_l$. $G_l$ is partitioned into nonoverlapping squares of size $S\times S$ and the $S^2$ nodes (i.e., pixels or superpixels) $p_l^{(s)} (s=1,\dots,S^2)$ in each square are merged into one bigger node (superpixel) $p_{l+1}$ in $G_{l+1}$, whose intensity is calculated as the average intensity in the square:
$$I(p_{l+1})=\frac{1}{S^2} \sum_{s=1}^{S^2} I(p_l^{(s)})$$
where $I(p)$ denotes the intensity of pixel (or superpixel) $p$. Let $W_l$ and $H_l$ be the width and height of the image in $G_l$, respectively, $d_l$ be the maximum disparity in $G_l$, and $d_0 = d$ is the maximum disparity of the original image.
Then, we have $W_{l+1} = \lceil W_l/S\rceil, H_{l+1} = \lceil H_l/S\rceil, |V_{l+1}| = H_{l+1}\times W_{l+1} \approx \lfloor|V_l|/S^2\rfloor,d_{l+1} = \lfloor d_l/S\rfloor$, where $\lceil \cdot \rceil$ and $\lfloor \cdot \rfloor$ are the ceiling function and floor function, respectively, $|A|$ denotes the number of elements in set $A$.
Our approach proceeds in a top-down manner. We want to predict the disparity range of layer $l$ from its upper layer $l+1$.
Let $D_l$ be the random variable of disparity in layer $l$, some key observations from the graph pyramid include: 
\begin{enumerate}
 \item \label{obs:sgl} We experimentally observed that $\forall l, 0\leq l\leq L-1$,
	the conditional probability $P(D_{l+1}|D_l)$ follows some regular pattern:
	If the disparity $D_l$ in layer $G_l$ is $\delta$, then the corresponding $D_{l+1}$
	in layer $G_{l+1}$ densely concentrates around $\delta/S$, as illustrated in Fig.~\ref{fig:sgl}.
	This is because if in $G_l$ pixel $p_l$ in the left image with coordinates $(x, y)$ corresponds to $(x-\delta, y)$
	in the right image, then in $G_{l+1}$, their corresponding coordinates become
	$(\lfloor  x/S\rfloor,\lfloor y/S\rfloor )$ and $(\lfloor(x-\delta)/S\rfloor,\lfloor y/S\rfloor )$.
 \item \label{obs:similarity}
	The conditional probabilities $P(D_{l+1}|D_l=j)$ for different $j$ are very similar. Specifically, $\forall j, 0\leq j\leq d_l$, the distributions $P(D_{l+1}-\lfloor j/S\rfloor|D_l=j)$ are very similar, where $\lfloor j/S\rfloor$ is the mode, see Fig.~\ref{fig:sgl}. It can be seen that these distributions can be well modeled by Gaussian Mixture Model (GMM).
\item \label{obs:lgs}
	Given the disparity of pixel $D_{l+1}^{(p)}$ in the higher layer, the possible range of $p_l^{(s)}$'s disparity $D_l^{(p^{(s)})}$ in layer $l$ can be faithfully predicted. Fig.~\ref{fig:lgs} (column 2) depicts the probability distributions $P(D_l|D_{l+1})\ (l=0,1)$. We found that the probability is large only within a small area around the diagonal direction. 
\end{enumerate}

These observations reveal that the disparities of the consecutive layers are closely  correlated. This motivates us to develop an algorithm that first computes disparities in a higher layer $l+1$ (which is faster and more robust), then predicts disparities in a lower layer $l$. Thus, the huge computation effort in the lower layer can be dramatically reduced. Also, these observations  motivate us to propose an Bayesian prediction model \cite{GMM_Bayes} for our HDP.

\subsection{Hierarchical Disparity Prediction Model}
In this subsection, we present out hierarchical disparity prediction model (HDPM). We compute the conditional probability of $D_l$ given $D_{l+1}$ by the Bayes' theorem as 
\begin{equation}
P(D_l|D_{l+1}) \propto P(D_{l+1}|D_l)P(D_l), 0\leq l\leq L-1, 
\end{equation}
where $P(D_{l+1}|D_l)$ can be modeled by GMM and $P(D_l)$ is approximated by sampling approach. Since the disparity takes discrete integer values, the probability can be written in matrix form,
\begin{multline}
\label{eq:dpm}
\mathbf{P}(\mathbf{D}_l|\mathbf{D}_{l+1}) \propto \mathbf{P}(\mathbf{D}_{l+1}|\mathbf{D}_l)\mathbf{P}(\mathbf{D}_l), 0\leq l\leq L-1
\end{multline}
where the bold letter denotes matrices, the $(i,j)$-th elements of the first two matrices are defined as $P(D_l=j|D_{l+1}=i)$ and $ P(D_{l+1}=i|D_l=j)$, respectivelty, and the thrid matrix is 
\begin{align*}
\mathbf{P}(\mathbf{D}_l) = {\rm{diag}} \Big(P(D_l=0), P(D_l=1),\cdots,P(D_l=d_{l})\Big).
\end{align*}

Now we address the problem of modeling $P(D_{l+1}|D_l)$ by GMM. Consider the conditional probability 
$P(D_{l+1}^{(p)} = \lfloor j/S\rfloor+o|D_l^{(p^{(s)})} = j)$, where $p^{(s)}$ is a pixel at level $l$, $p$ is a pixel at level $l+1$, and $o$ is assumed to be the offset of $D_{l+1}^{(p)}$ from the mode $\lfloor D_l^{(p^{(s)})}/S\rfloor$.  
From observation \ref{obs:similarity}, we can assume that the two events
$D_{l+1}^{(p)}-\lfloor D_l^{(p^{(s)})}/S\rfloor=o$ and $D_l^{(p^{(s)})}=j$
are independent. So we have 
\begin{multline}
P(D_{l+1}^{(p)} = \lfloor j/S\rfloor+o|D_l^{(p^{(s)})} = j) \\
= P(D_{l+1}^{(p)}-\lfloor D_l^{(p^{(s)})}/S\rfloor=o).
\end{multline}
Therefore, if we can obtain the distribution vector
\begin{multline*}
P(\mathbf{D}_{l+1}-\lfloor \mathbf{D}_l/S\rfloor)
=[P(D_{l+1}-\lfloor D_l/S\rfloor=-d_l), \\ P(D_{l+1}-\lfloor D_l/S\rfloor 
=-d_l+1),\dots,P(D_{l+1}-\lfloor D_l/S\rfloor=d_l)]^T
\end{multline*}
and model it by a GMM, then we can shift the vector of GMM distribution and concatenate them to obtain the matrix $\mathbf{P}(\mathbf{D}_{l+1}|\mathbf{D}_l)$.  
In our experiment later, we randomly selected half of images from the Middleburry 2006 dataset \cite{Midd4} to form training dataset to train a GMM. After training by EM \cite{GMM_Bayes}, we obtain the parameters: the mean vectors $\mu_l$, the standard deviation $\sigma_l$ and the mixing coefficients $\pi_l$. Therefore, the GMM distribution for $G_l$, ${\bf GMM}_l$, is defined as
\begin{multline*}
P(\mathbf{D}_{l+1}-\lfloor \mathbf{D}_l/S\rfloor=o) \sim {\bf GMM}_l(o)=\\
\sum_{k=1}^{K_l}\mathbf{\pi_l}(k)\mathcal{N}(o|\mathbf{\mu}_l(k),\mathbf{\sigma}_l(k)).
\end{multline*}
From this, we can generate the matrix $\mathbf{P}(\mathbf{D}_{l+1}|\mathbf{D}_l)$ whose $(i,j)$-th element is 
\begin{equation}
P(D_{l+1}=i|D_l=j)={\bf GMM}_l(i-\lfloor j/S \rfloor).
\end{equation}

\section{Stereo Matching based on HDP }

\begin{figure}
	\centerline{\includegraphics[width = 8.5cm]{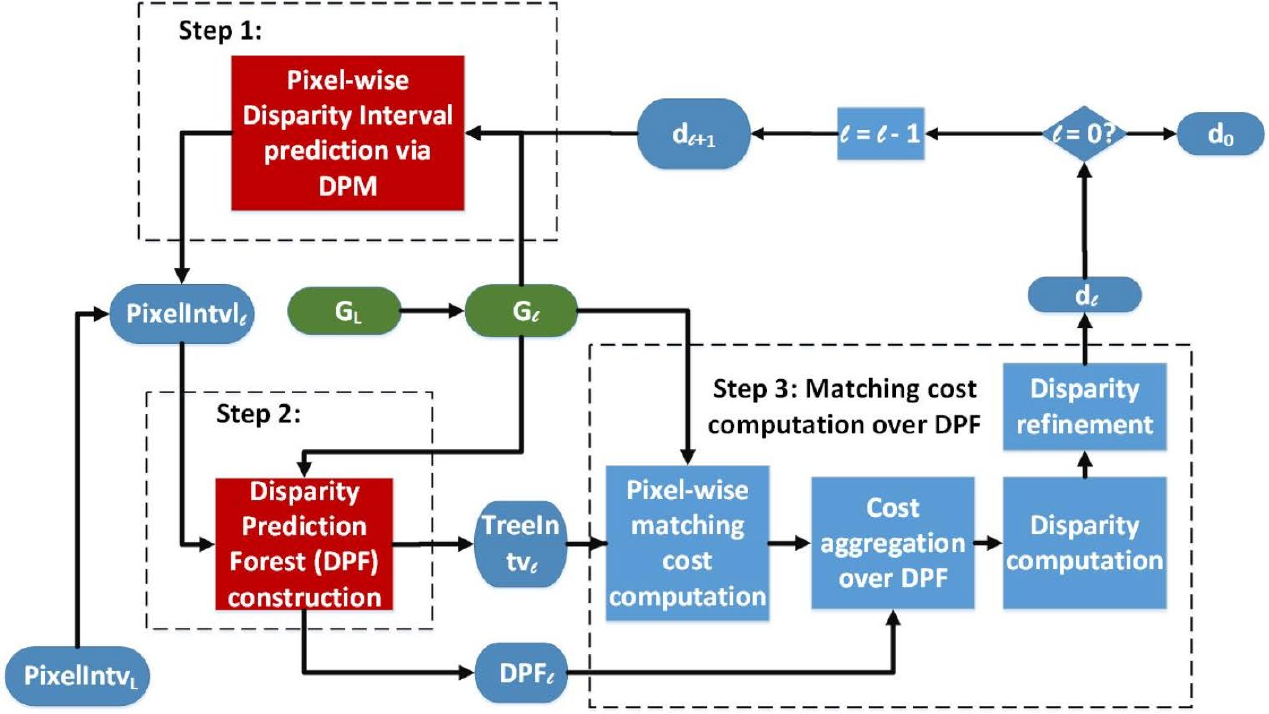}}
	\caption {Overall procedure of the proposed HDP framework. $G_l$ and $d_l$   represent the input graph and the output disparity map at layer $l$.}
	\label{fig:overallprocess}
\end{figure}

In this section, we integrate our HDP to the tree-based stereo mathching methods. Fig.\ref{fig:overallprocess} illustrates the overall procedure of the proposed hierarchical aggregation strategy. 
The process goes top-down from layer $G_L$ to $G_0$ and the final disparity is obtained from layer $G_0$ (the input image itself). The procedure in each layer can be summarized into three main steps: 1) pixel-wise disparity interval prediction; 2) disparity prediction forest (HDPF) construction; 3) matching cost aggregation over HDPF. 

\subsection{Pixel-wise Disparity Interval Prediction}
This step focuses on predicting a smaller disparity interval $PixelIntv$ for each pixel using our HDPM. 
Initially, the disparity interval for each pixel $p_L$ at the highest layer $G_L$ is set to be $[0,d_L]$, $d_L=\lfloor d/S^L \rfloor $. 
For each layer $0 \leq l \leq L-1$, we need to compute $\mathbf{P(D}_l)$ and $\mathbf{P}(\mathbf{D}_{l+1}|\mathbf{D}_l)$, and then calculate $\mathbf{P}(\mathbf{D}_l|\mathbf{D}_{l+1})$ using Eq.~(\ref{eq:dpm}).
To compute $\mathbf{P(D}_l)$, we sample a set of pixels and calculate their disparities to obtain the approximate distribution. Specifically, we partition $G_l$ into $5\times 5$ nonoverlapping squares, and use the method in \cite{Refinement} to calculate the disparity for the center of each square and find the stable pixels by left-to-right consistency check \cite{Refinement}. Since the disparities of stable pixels are reliable, we approximate disparity distribution $\mathbf{P(D}_l)$ of all pixels in $V_l$ by the disparity distribution of these pixels.  $\mathbf{P}(\mathbf{D}_{l+1}|\mathbf{D}_l)$ can be modeled by ${\bf GMM}_l$ as described above. Finally, we normalize each row of $\mathbf{P}(\mathbf{D}_{l+1}|\mathbf{D}_l)\mathbf{P(D}_l)$ to get $\mathbf{P}(\mathbf{D}_l|\mathbf{D}_{l+1})$ as shown in Eq.~(\ref{eq:dpm}). 

Next, given $D_{l+1}^{(p)}=i$, we use $\mathbf{P}(\mathbf{D}_l|\mathbf{D}_{l+1})$ to predict a disparity interval for $D_l^{(p^{(s)})}$, denoted as $PixelIntv(\{p_l^{(s)}|D_{l+1}^{(p)}=i\})$. The basic idea is to first set $PixelIntv(\{p_l^{(s)}|D_{l+1}^{(p)}=i\})$ to be the column index with the highest probability in the $i$th row of the matrix $\mathbf{P}(\mathbf{D}_l|\mathbf{D}_{l+1})$, i.e., $PixelIntv(\{p_l^{(s)}|D_{l+1}^{(p)}=i\}) = \arg\max_{0\leq j\leq d_{l}} \mathbf{P}(i,j)$ where $\mathbf{P}(i,j)$ denotes the $(i,j)$-th element in the matrix $\mathbf{P}(\mathbf{D}_l|\mathbf{D}_{l+1})$. 
Then we extend $PixelIntv(\{p_l^{(s)}|D_{l+1}^{(p)}=i\})$ to $PixelIntv(\{p_l^{(s)}|D_{l+1}^{(p)}=i\}) \cup \{j\} $ if $j$ satisfies the following criterion:
\begin{equation} \label{eq:crtr}
\frac{P(D_l=j|D_{l+1}=i)}{c+P(D_l=j|D_{l+1}=i)} \geq \delta_l
\end{equation}
where $\delta_l$ is a threshold parameter and $c = \sum_j \mathbf{P}(i,j)$ for $j \in PixelIntv(\{p_l^{(s)}|D_{l+1}^{(p)}=i\})$ is the sum of the probabilities of disparities that have been selected so far. We can see that when $c$ is small at the begining, an index is easy to be added to the disparity interval, but as $c$ becomes larger, an index becomes harder to be added.

\subsection{HDP Forest Construction and Tree-wise Disparity Interval Calculation}

To enable aggregation over pixels with similar disparity intervals, we need to merge pixels with similar disparity intervals into subtrees as shown in fig.~\ref{fig:forest-segment} (Each different color represents a different subtree). Such a subtree is called a \textbf{Disparity Tree (DT)}. Therefore, we are actually performing depth-based segmentation. Each DT represents a region with similar disparities. In each layer $l$, we use $TreeIntv(DT_{l,i})$ to denote tree-wise disparity interval of $DT_{l,i}$, where $DT_{l,i}$ is the $i$th DT in layer $l$. The tree-wise disparity interval for $DT_{l,i}$ is simply defined as the union of the pixel-wise disparity intervals of the pixels in $DT_{l,i}$. These tree-wise disparity intervals are exactly the ones we use to replace the full allowed disparity range in search of disparity. There are two advantages of using DT and tree-wise disparity intervals instead of pixel-wise ones: 
1) Pixels in the same DT share the same tree-wise disparity intervals, and therefore, their cost aggregation can be carried out together over the same disparity candidates.
2) Pixels in different DT's (and thus having dissimilar disparities) are supposed to be disconnected in the graph, which guarantees no impact between them when we aggregate matching cost. We call the set of independent DT's 
$HDPF_l = \{DT_{l,1},\  DT_{l,2},\ \dots\ DT_{l,n_l}\}$ as the hierarchical disparity prediction forest (HDPF) in layer $l$. 

In a general spanning tree 
algorithm, there are two rules, called \textbf{Rule 1} and \textbf{Rule 2}. \textbf{Rule 1} is the rule to pick an edge from all the candidates in the input graph $G$ (e.g., pick the edge of the minimum weight from the rest edges in MST). \textbf{Rule 2} decides whether an edge should be added in the spanning tree (e.g., the two ends of the edge are not already in the same tree in MST). 

Our HDPF construction is implemented by adding two rules: \textbf{DT-Rule 1} and \textbf{DT-Rule 2}, which take disparity information into account when we build HDPF. Our HDPF construction and tree-wise disparity interval calculation algorithm proceeds in three stages:

\textbf{Initialization}: Calculate the weight of the edges $e(p_l, q_l)$ in $E_l$ by
\begin{equation}
\label{eq:weight}
w(p_l, q_l) = \left|I(p_l) - I(q_l)\right|,
\end{equation}
where $I(p_l)$ and $I(q_l)$ represent the intensity of the pixel $p_l$ and $q_l$ in $G_l$ as in \cite{MST}.
An initial disparity tree $DT_l^{(p)}$ is created and the tree-wise disparity interval is initialized as the pixel-wise disparity interval at that pixel. 

\textbf{Selecting edges}: We apply \textbf{DT-Rule 1} to remove undesired edges selected by \textbf{Rule 1}. DT-Rule 1 is stated as: for two neighboring pixels $p_l$ and $q_l$, if
\begin{equation}
\label{eq:cri1}
PixelIntv(p_l) \cap PixelIntv(q_l) = \emptyset, p_l, q_l \in V_l.
\end{equation}
then the edge connecting them will be discarded.

\textbf{Merging trees}: Two DTs are merged into a bigger DT if both \textbf{Rule 2} and \textbf{DT-Rule 2} are satisfied. \textbf{DT-Rule 2} is stated as: for two DTs $DT_l^{(p)}$ and $DT_l^{(q)}$, if 
\begin{equation}
\label{eq:DTrules}
\frac{|TreeIntv(DT_l^{(p)}) \cap TreeIntv(DT_l^{(q)})|}{|TreeIntv(DT_l^{(p)}) \cup TreeIntv(DT_l^{(q)})|} \ge \beta,
\end{equation}
where $\beta$ is a threshold parameter, then they are merged into a new DT $DT_l^{(p,q)}$. And at the same time, the $TreeIntv$ is updated by the union operation, i.e., 
\begin{equation}
TreeIntv(DT_l^{(p,q)})=TreeIntv(DT_l^{p})\cup TreeIntv(DT_l^q) 
\end{equation}

\subsection{Matching Cost Aggregation Over HDPF}
\label{sec:agg}
Like many other popular cost aggregation strategies, this step goes into four stages: 1) pixel-wise matching cost computation; 2) cost aggregation; 3) disparity computation and 4) disparity refinement. However, unlike existing methods, we implement the first three stages over the generated HDPF within each DT independently. 

In our method, the pixel-wise matching cost will be hierarchially refined layer by layer according to the generated HDPF and the tree-wise disparity intervals. The new matching cost for each pixel $p_l$ at disparity $x\in [0,d_l]$ in layer $G_l$ can be expressed as:
\begin{equation}
\label{eq:E}
E\left(p_l,\ x\right) =
\begin{cases}
M\left(p_l,\ x\right) & \text { if } x \in TreeIntv(DT_l^{(p)}) \\
undefined & \text {otherwise}
\end{cases}
\end{equation}
were $M(p_l, x)$ is the pixel-based matching cost proposed in \cite{Cost2}. It calculates the truncated absolute color difference in terms of RGB and the gradient in horizontal direction at the matching points.

As is shown in Eq.(\ref{eq:E}), the matching cost will not be updated for all the disparity candidates, but only for those that fall into the corresponding tree-wise disparity interval of the pixels. 
The length of the path from $p_l$ to $q_l$ along $HDPF_l$ is defined as:
\begin{equation}
\label{eq:W}
W\left(p_l,\ q_l \right) = \sum_{e\left(s_l,\ r_l\right) \in path\left(p_l,\ q_l\right)}w\left(s_l,\ r_l\right),
\end{equation}
where $p_l$ and $q_l$ are in the same $DT$. The similarity $S(p_l, q_l)$ between pixels $p_l$ and $q_l$ is defined similarly as in \cite{MST}:
\begin{equation}
\label{eq:S}
S\left(p_l,\ q_l\right) =
\begin{cases}
e^{-\frac{W\left(p_l,\ q_l\right)}{\gamma}} & \text{ if } q_l \in DT_l^{(p)}  \\
0 & \text{ otherwise }
\end{cases},
\end{equation}
where $\gamma$ is a constant to adjust the similarity.

The aggregated cost $C(p_l, x)$ for each pixel $p_l$ in layer $G_l$ at disparity label $x$ is:
\begin{equation}
\label{eq:C}
\begin{split}
C\left(p_l,\ x\right) = \sum_{q_l \in DT_l^{(p)}}S\left(p_l,\ q_l\right) \cdot E\left(q_l,\ x\right), \\
x \in TreeIntv(DT_l^{(p)}).
\end{split}
\end{equation}
Obviously, the similarity is 0 if two pixels are in the different DTs. 
This is because there would be no path to connect pixels which are separated into different disparity regions and such pixels should have no impact on each other in cost aggregation. Hence, the cost aggregation happens in each DT (support region) independently. Finally, the disparity label that minimizes the overall aggregated cost for each pixel is thus chosen as its disparity, this is the so-called winner-takes-all scheme. 

To avoid the effect of different refinement methods, we did not apply refinement methods in our experiments in the next section. However, the tree-based refinement methods, as proposed in \cite{MST,ST}, can be easily integrated into our framework to further improve the accuracy of initial disparity at each layer.

 \section{Experiments}
 \label{sec:experiment}

 \subsection{Datasets and Experiment settings}

In this section, we evaluate the performance of the proposed hierarchical disparity prediction model on three datasets: Middlebury \cite{Midd4}, KITTI \cite{KITTI} and a low-quality image dataset we collected, which include both laboratory, real-world and low-quality images.

We combine our hierarchical disparity prediction model with the MST \cite{MST}, ST \cite{ST} algorithms. We also present a random spanning tree (RT) algorithm, where we randomly shuffle the edges and use the union-find algorithm to generate a random spanning tree over which the cost is aggregated. We find that sometimes RT outperforms MST and ST. We discover that MST \cite{MST} and ST \cite{ST} both apply a median filter before building their trees, which degrades their accuracy. For fair comparison, we remove the median filter in ST, MST, RT, and their HDP versions (HDP+MST, HDP+ST and HDP+RT) and denote the versions with the filter as M+*.

\textbf{Middlebury}:
Middlebury is the most widely-used dataset for stereo matching.
We select Middelbury 2006  datasets \cite{Midd4} which contain overall 21 test images with all ground truth disparity maps available in three resolutions:
third size (width: 443$\sim$463, height: 370), half size (width: 665$\sim$695, height: 555) and full size (width:1240$\sim$1396, height:1110).

\textbf{KITTI}:
The KITTI dataset \cite{KITTI} is a new dataset captured by driving around Karlsruhe, in its rural areas and on highways, and is used in the mobile robotics and autonomous driving research.
In our experiments, we select a large KITTI subset (453 frames, each of resolution $1242 \times 375$) of the  ``$2011\_09\_26\_drive\_0009$" sequence that contains mostly ``car'' objects.

\textbf{Low-quality dataset}:
We capture our own datasets within an urban area using two regular web cameras (resolution: $640 \times 480$). 
Illumination deviation exists in these image pairs due to environment conditions (e.g. sunlight).
Non-textured regions occupy more than half of the image for some test cases. 267 image pairs from more than twenty different scene categories are selected to evaluate performance The images are pre-processed only by rectification.

Throughout the experiments, we set $S = 2, L = 3, \delta_l = \delta_0 S^l$. For large images in KITTI \cite{KITTI} and Middlebury's full-size images \cite{Midd4}, we set $\delta_0 = 0.004, \beta = 0.95$. For small images in Low-quality dataset and half-size images \cite{Midd4}, $\delta_0=0.064, \beta = 0.6$.
 	
\subsection{Computational Efficiency Comparisons}
We define the ratio between the length of average disparity search range using HDP and $G_l$'s complete disparity range $[0,d_l]$ as 
$	R_l = \frac{1}{|V_l|} \sum_{p_l\in V_l} \frac{|TreeIntv(DT^{(p)}_l)|}{1+d_l}$
to measure the computational efficiency of our HDPM. The results are presented in table \ref{tab:avgsearchratio}, we can see that the average search ratio is narrowed down layer by layer. On average, HDPM can reduce the search range to about one-tenth of the original algorithms for half-size images and about one-twentieth for the full-size images.

\begin{table}
	\centering
	\begin{tabular}{p{2cm}p{1.5cm}p{1.5cm}p{1.5cm}} 
		\hline
		layer & 0 & 1 & 2 \\ \hline
		{Half size} & {10.3} & {11.1} & {100.0} \\ 
		{Full size} & {1.5} & {2.7} & {100.0}\\ \hline
	\end{tabular}
	\caption{HDP+ST's $R_l$ ($0\le l < L$) of testcases in Middlebury 2006 dataset \cite{Midd4} of half-size and full-size resolutions.}
	\label{tab:avgsearchratio}
\end{table}

\begin{figure}
 	\centerline{\includegraphics[width = \linewidth]{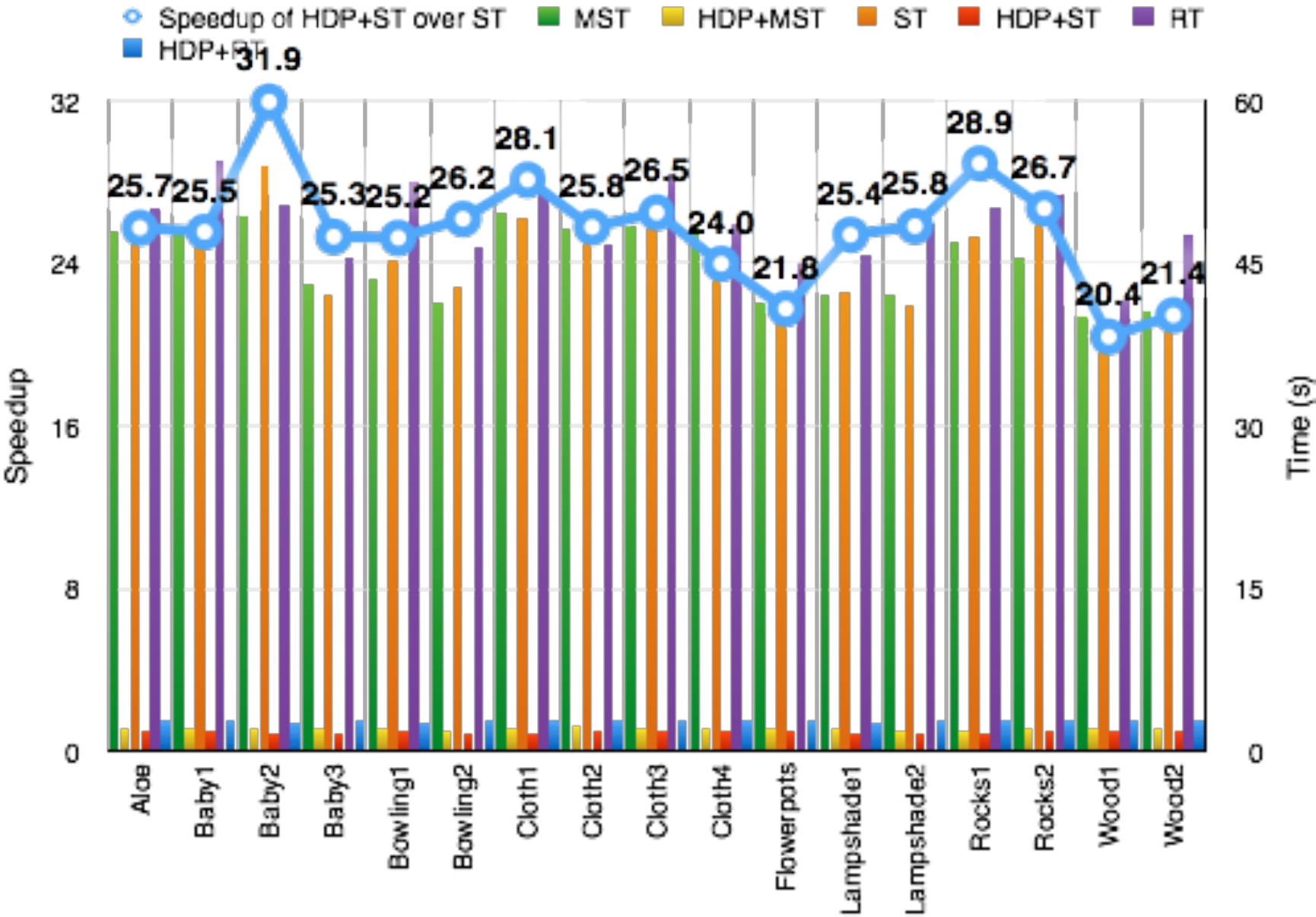}}
 	\caption {Comparison of runtime on Middlebury 2006 dataset \cite{Midd4} of full-size resolution: MST, ST and RT vs. their HDP versions.}
 	\label{fig:runtime}
 \end{figure}

The runtime of MST, ST, RT and their HDP versions 
  is shown in Fig.\ref{fig:runtime}, where bars represent the time (in seconds) and circles represent the speedup of HDP+ST over ST. The average runtime of MST, ST and RT is all above 45s, but the runtime of their HDP versions is about 2s. The average speedup is 25.57 for HDP+ST over ST. The average speedups of HDP+MST and HDP+RT are 20.96 and 17.12, respectively.
 
 \subsection{Performance Comparisons}
 This subsection quantitatively evaluates the accuracy of the three tree-based algorithms and their HDP versions. 
 \paragraph{\bf Performance on Middlebury}
The algorithms are tested on 17 pairs of half-size images and 17 pairs of full-size images from the Middlebury 2006 dataset \cite{Midd4}. Following standard practice, we evaluate the error rate in non-occluded regions under two metrics: err $\ge$ 1 and err $\ge$ 2 consider a pixel erroreous if its estimated disparity differs from the ground truth by greater than or equal to 1 or 2  pixels, respectively. Table \ref{tab:midd-result} compares the error rate of each tree-based algorithm with its HDP and M+* versions on the 17 full-size images under err $\ge$ 2 metrics. Table \ref{tab:avgerr} summarizes their average error rate under both metrics on both half-size and full-size images. Each HDP version achieves lower average error rate than the original one. Especially, HDP versions' error rates are is 2.2\%, 2.2\% and 4.9\% (err $\ge$ 2) lower than MST, ST and RT on full-size images. Also note that HDP+RT achieves the lowest error on both half-size and full-size images under err $\ge$ 2 metric.

\begin{table}
 	\centering
	\begin{tabular}{m{0.8cm}|m{0.6cm}m{0.4cm}m{0.5cm}|m{0.4cm}m{0.3cm}m{0.4cm}|m{0.4cm}m{0.3cm}m{0.4cm}}
		\hline
		\multicolumn{10}{c}{Non-occluded error rate (\%) (error $\geq$ 2.0) } \\
		\hline
		 &
		HDP +MST &
		MST &
		M+ MST &
		HDP +ST &
		ST &
		M+ ST &
		HDP +RT &
		RT &
		M+ RT \\
		\hline
		{Aloe} & {7.5} & {\bf 4.6} & {5.4} & {7.3} & {\bf 4.3} & {4.9} & {10.0} & {5.6} & {\bf 5.6} \\ \hline
		{Baby1} & {\bf 8.1} & {11.5} & {10.0} & {\bf 7.7} & {11.5} & {10.4} & {\bf 7.8} & {17.4} & {14.8} \\ \hline
		{Baby2} & {\bf 22.5} & {25.9} & {29.6} & {\bf 20.9} & {25.3} & {28.4} & {\bf 16.0} & {27.2} & {24.5} \\ \hline
		{Baby3} & {\bf 9.2} & {10.4} & {9.4} & {9.4} & {10.5} & {\bf 9.2} & {\bf 10.2} & {14.6} & {12.4} \\ \hline
		{Bowl1} & {\bf 19.8} & {29.8} & {33.0} & {\bf 20.4} & {29.7} & {32.7} & {\bf 20.7} & {36.3} & {30.7} \\ \hline
		{Bowl2} & {\bf 16.2} & {18.2} & {20.4} & {\bf 15.6} & {17.9} & {19.9} & {\bf 14.3} & {17.8} & {17.1} \\ \hline
		{Cloth1} & {0.6} & {\bf 0.5} & {0.8} & {0.7} & {\bf 0.4} & {0.6} & {1.3} & {1.1} & {\bf 0.9} \\ \hline
		{Cloth2} & {6.1} & {\bf 6.0} & {7.1} & {5.7} & {\bf 5.4} & {6.4} & {\bf 5.2} & {6.6} & {6.5} \\ \hline
		{Cloth3} & {2.5} & {\bf 1.6} & {2.0} & {2.5} & {\bf 1.5} & {1.8} & {3.0} & {1.9} & {\bf 1.8} \\ \hline
		{Cloth4} & {2.6} & {\bf 2.1} & {2.3} & {2.3} & {\bf 2.0} & {2.1} & {2.5} & {2.6} & {\bf 2.4} \\ \hline
		{Flower} & {\bf 23.4} & {27.4} & {27.1} & {\bf 23.7} & {27.1} & {26.9} & {\bf 22.2} & {27.2} & {27.1} \\ \hline
		{Lamp1} & {\bf 14.2} & {16.4} & {18.1} & {\bf 14.3} & {17.3} & {18.1} & {\bf 16.3} & {28.9} & {23.5} \\ \hline
		{Lamp2} & {\bf 20.4} & {29.3} & {28.2} & {\bf 20.5} & {29.9} & {31.0} & {\bf 22.7} & {35.4} & {31.4} \\ \hline
		{Rocks1} & {\bf 6.5} & {7.9} & {7.8} & {\bf 6.7} & {7.5} & {7.3} & {\bf 7.2} & {9.5} & {8.2} \\ \hline
		{Rocks2} & {\bf 2.7} & {4.1} & {4.1} & {\bf 2.7} & {3.9} & {3.8} & {\bf 2.9} & {4.8} & {4.4} \\ \hline
		{Wood1} & {\bf 7.6} & {7.6} & {7.7} & {7.5} & {\bf 7.1} & {7.4} & {\bf 8.4} & {10.4} & {10.6} \\ \hline
		{Wood2} & {\bf 5.8} & {9.6} & {12.3} & {\bf 5.6} & {9.5} & {12.1} & {\bf 5.7} & {12.2} & {10.5} \\ \hline
		{Avg} & {\bf 10.3} & {12.5} & {13.3} & {\bf 10.2} & {12.4} & {13.1} & {\bf 10.4} & {15.3} & {13.7} \\ \hline

	\end{tabular}
 		\caption{Quantitative evaluation. (Bowl, Flower and Lamp, Avg are abbreviations for Bowling, Flowerpots and Lampshade, Average respectively.)
 		}
 		\label{tab:midd-result}
 	\end{table}

\begin{table}
	\centering
	\begin{tabular}{m{1.4cm}|m{1.3cm}m{1.3cm}|m{1.3cm}c}
		\hline
		& \multicolumn{2}{c}{Half Size (\%)} & \multicolumn{2}{|c}{Full Size (\%)} \\ \cline{2-5}
		&	err$\ge$ 1  &	err $\ge$ 2 &	err $\ge$ 1 & err $\ge$ 2 \\ \hline
		{HDP+MST} & {16.1} & {6.4} & {44.1} & {10.3} \\
		{MST} & {16.4} & {6.9} & {45.0} & {12.5} \\ 
		{M+MST} & {19.5} & {8.4} & {46.3} & {13.3} \\ \hline
		{HDP+ST} & {15.7} & {6.3} & {44.0} & {10.2} \\ 
		{ST} & {15.8} & {6.8} & {44.8} & {12.4} \\ 
		{M+ST} & {18.7} & {8.2} & {46.0} & {13.1} \\ \hline
		{HDP+RT} & {15.4} & {7.3} & {43.7} & {10.4} \\ 
		{RT} & {15.7} & {7.9} & {46.4} & {15.3} \\ 
		{M+RT} & {16.6} & {7.6} & {45.8} & {13.7} \\ \hline
		\end{tabular}
 	\caption{Average error rates.}
 	\label{tab:avgerr}
 \end{table}

\begin{figure}
	\begin{minipage}[b]{0.155\linewidth}
		\includegraphics[width = 1.35cm]{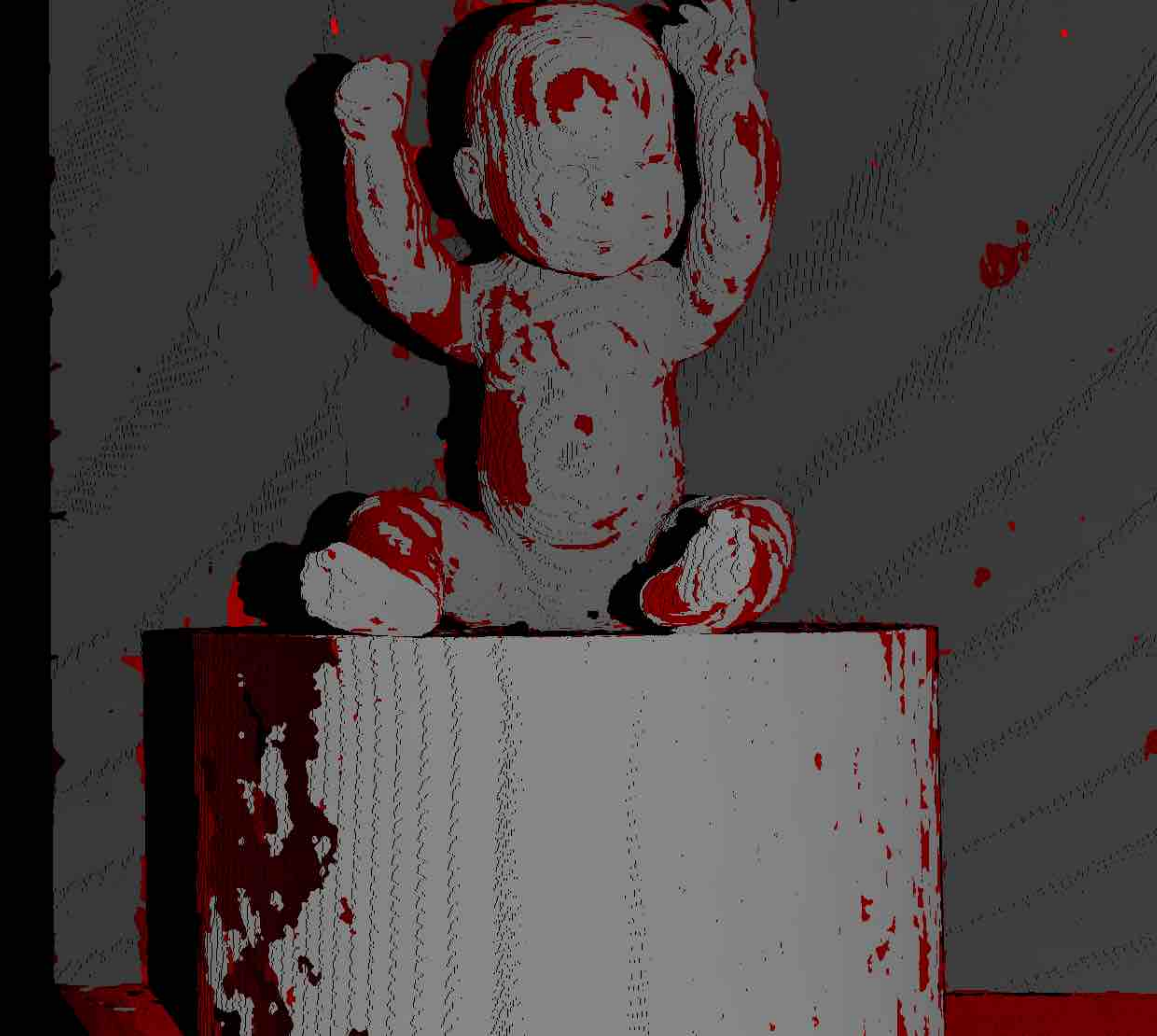}
	\end{minipage}
	\begin{minipage}[b]{0.155\linewidth}
		\includegraphics[width = 1.35cm]{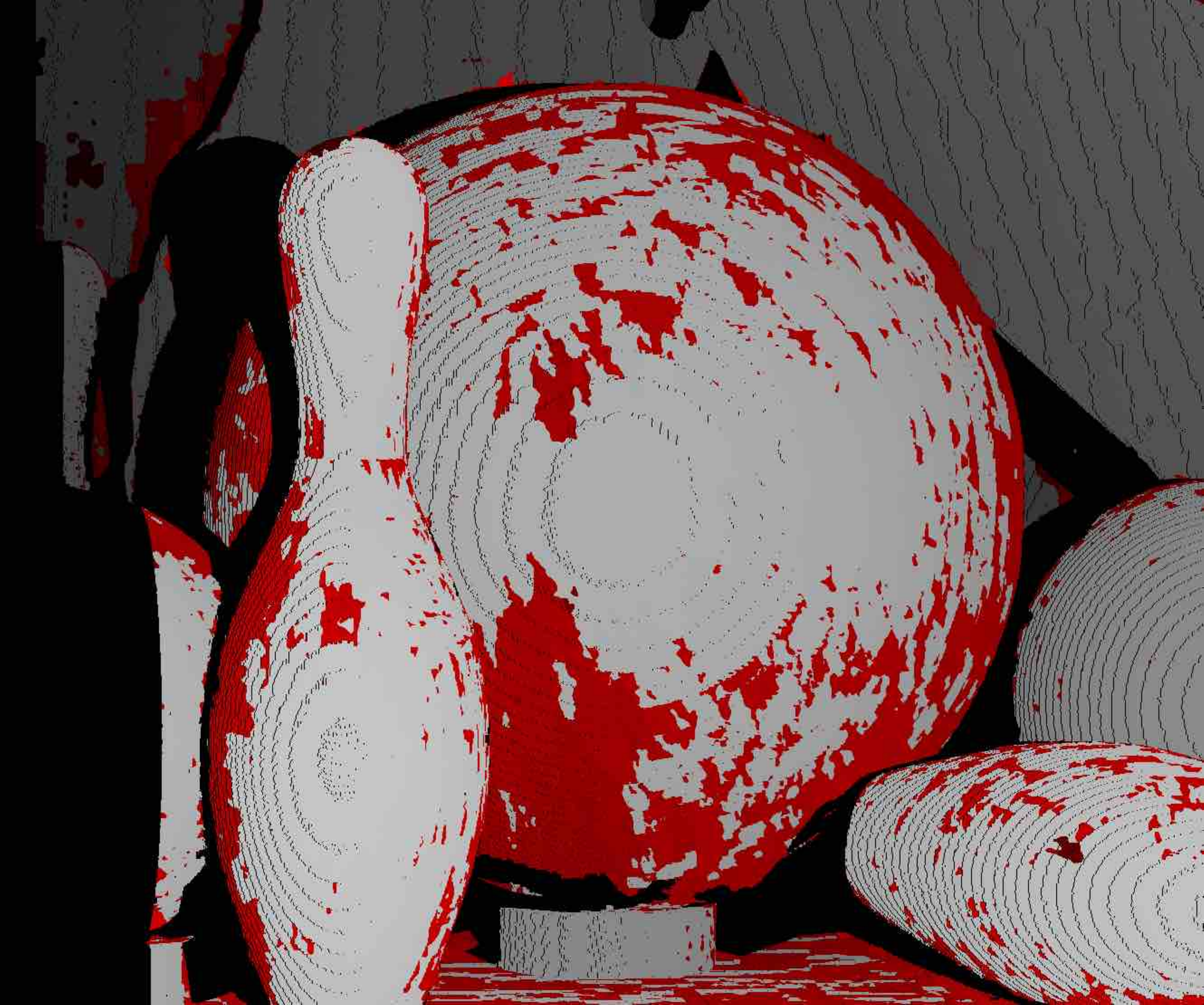}
	\end{minipage}
	\begin{minipage}[b]{0.17\linewidth}
		\includegraphics[width = 1.35cm]{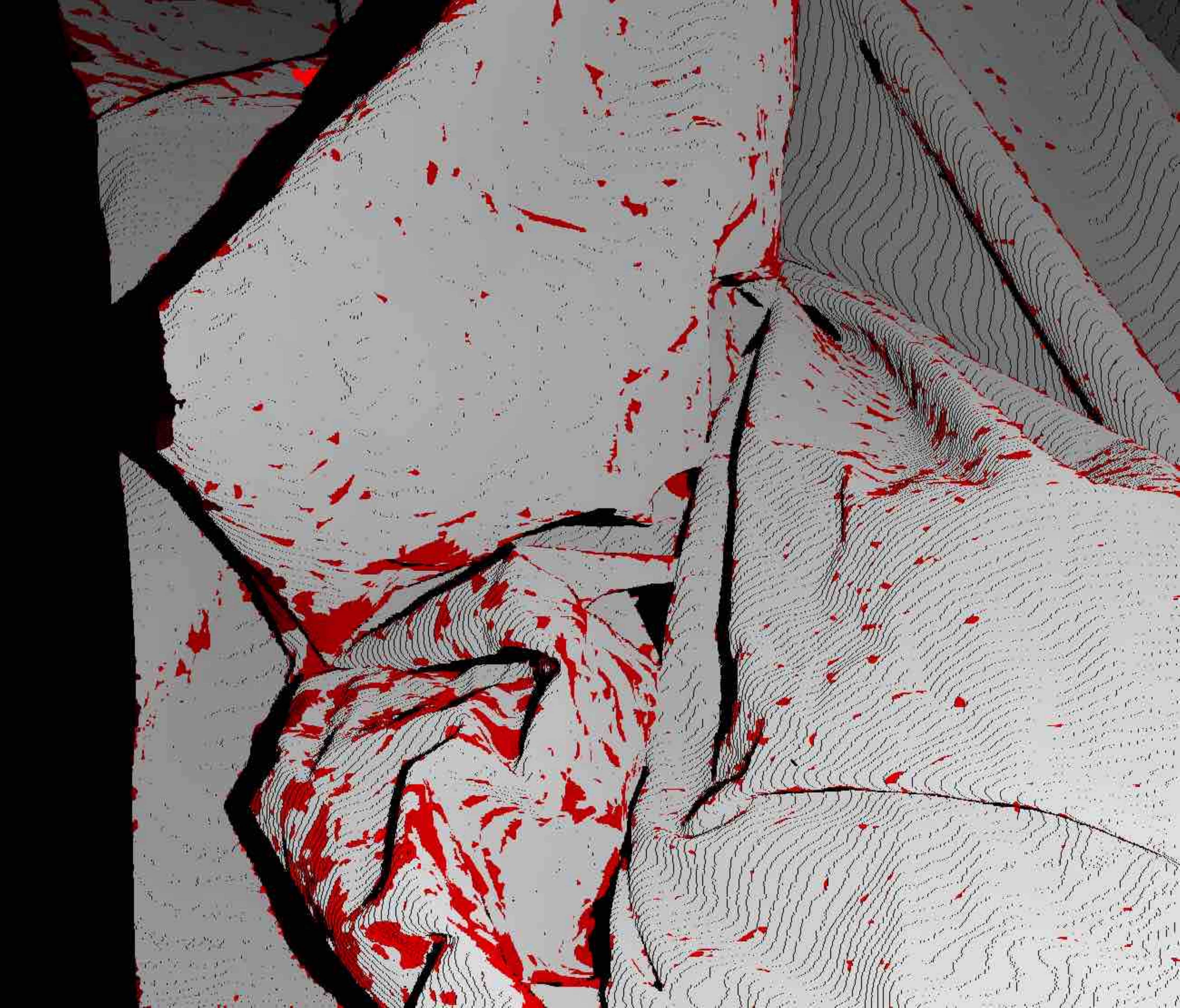}
	\end{minipage}
	\begin{minipage}[b]{0.155\linewidth}
		\includegraphics[width = 1.35cm]{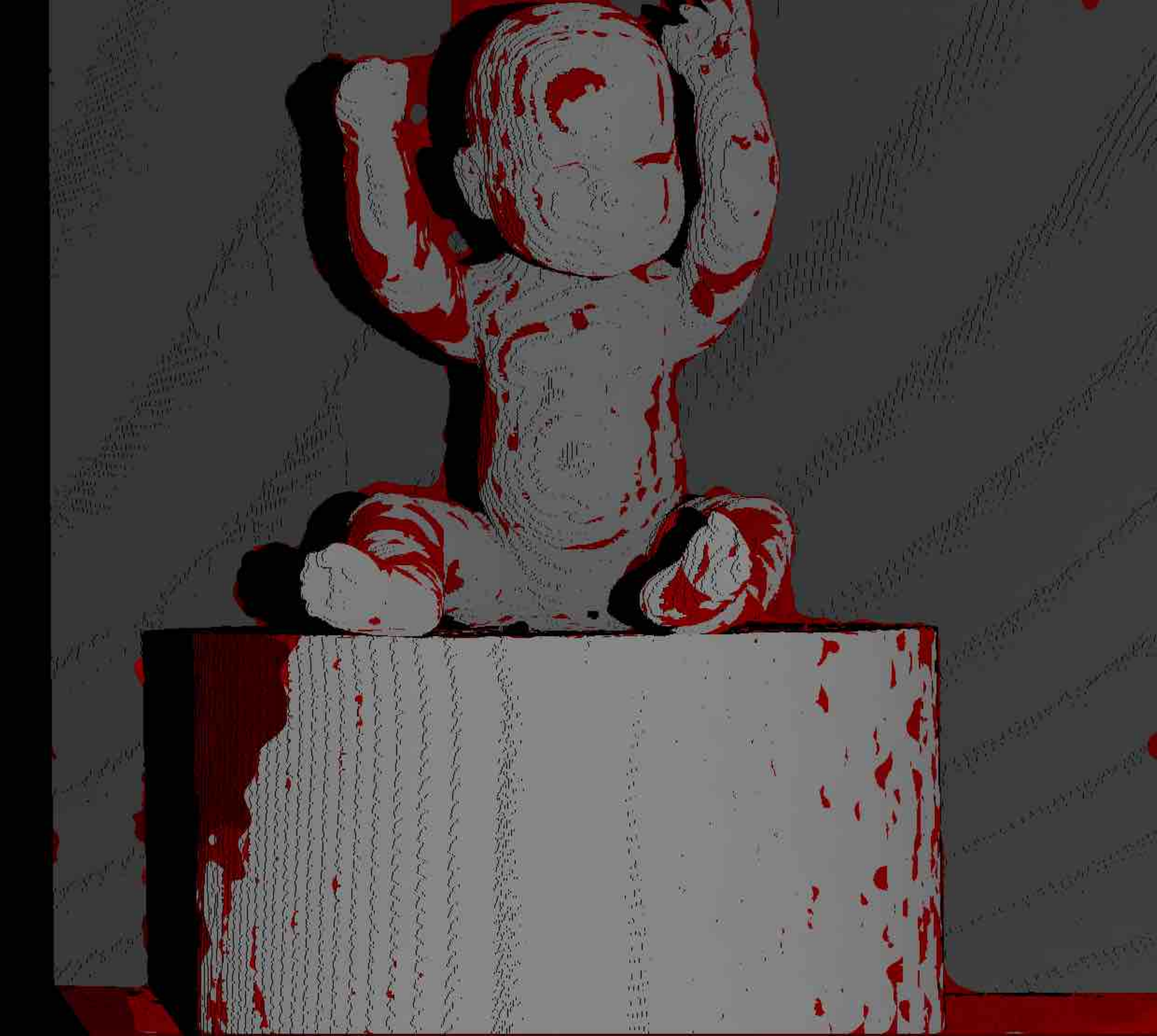}
	\end{minipage}
	\begin{minipage}[b]{0.155\linewidth}
		\includegraphics[width = 1.35cm]{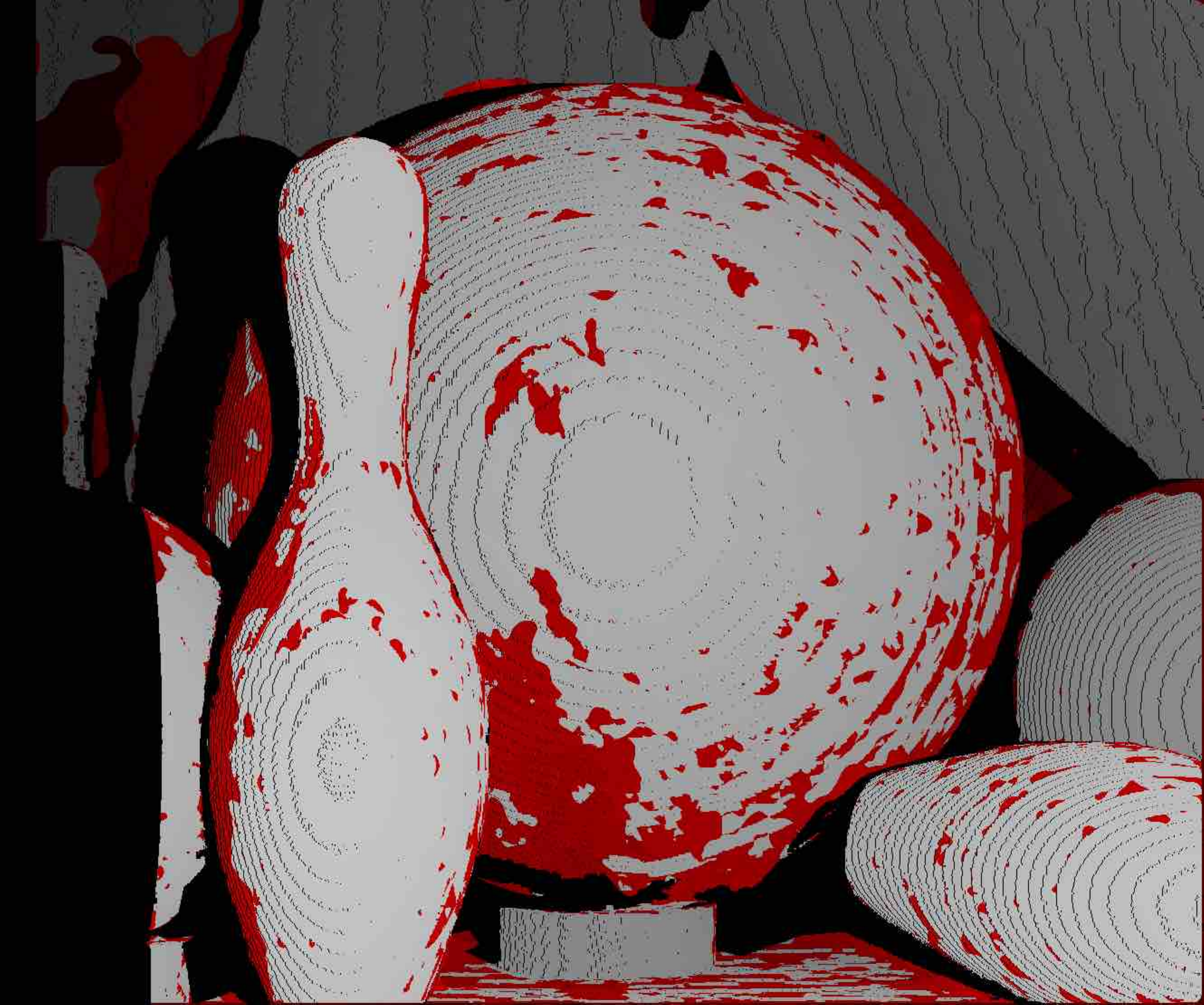}
	\end{minipage}
	\begin{minipage}[b]{0.155\linewidth}
		\includegraphics[width = 1.35cm]{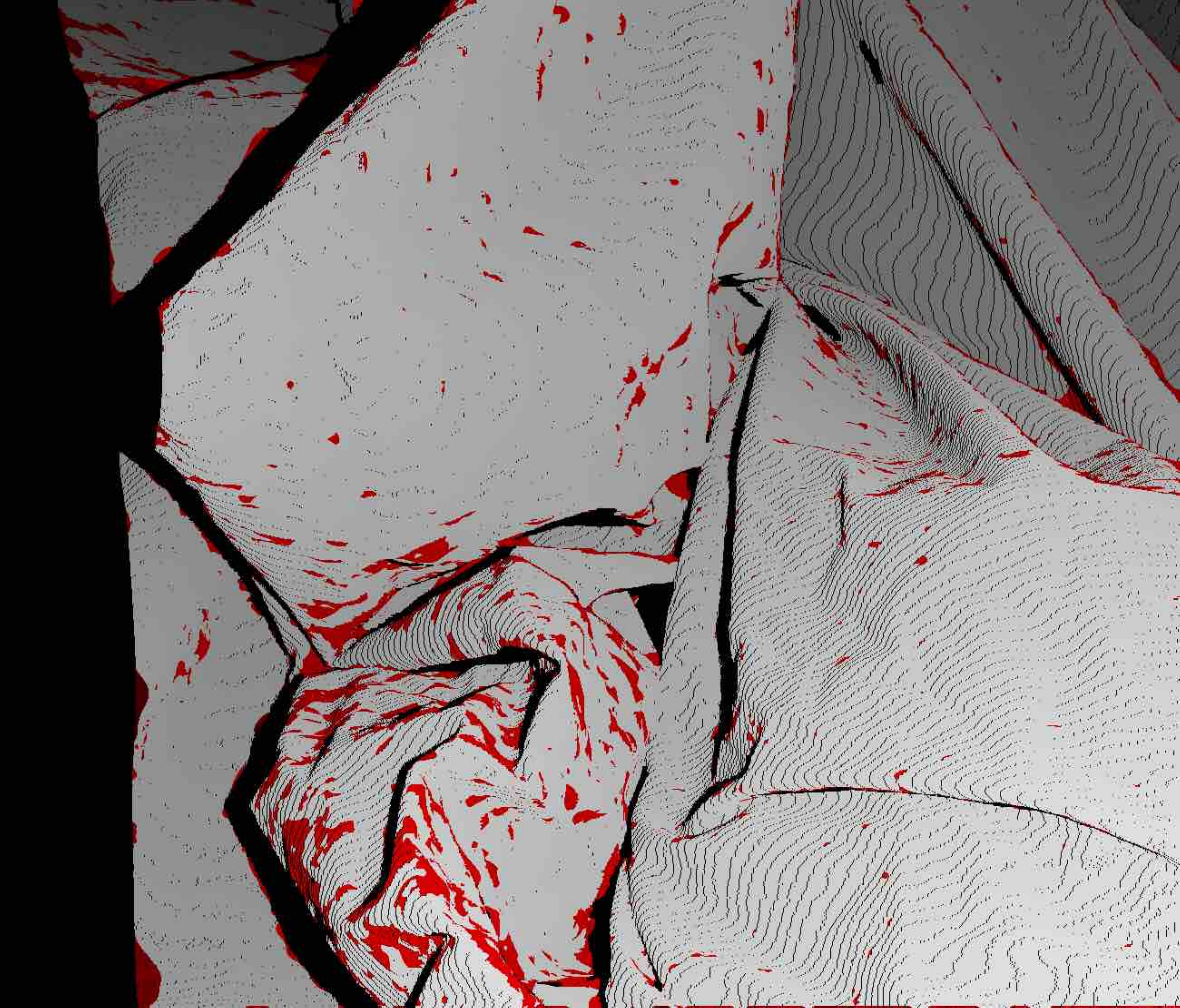}
	\end{minipage}

	\begin{minipage}[b]{0.155\linewidth}
		\center
		\includegraphics[width = 1.35cm]{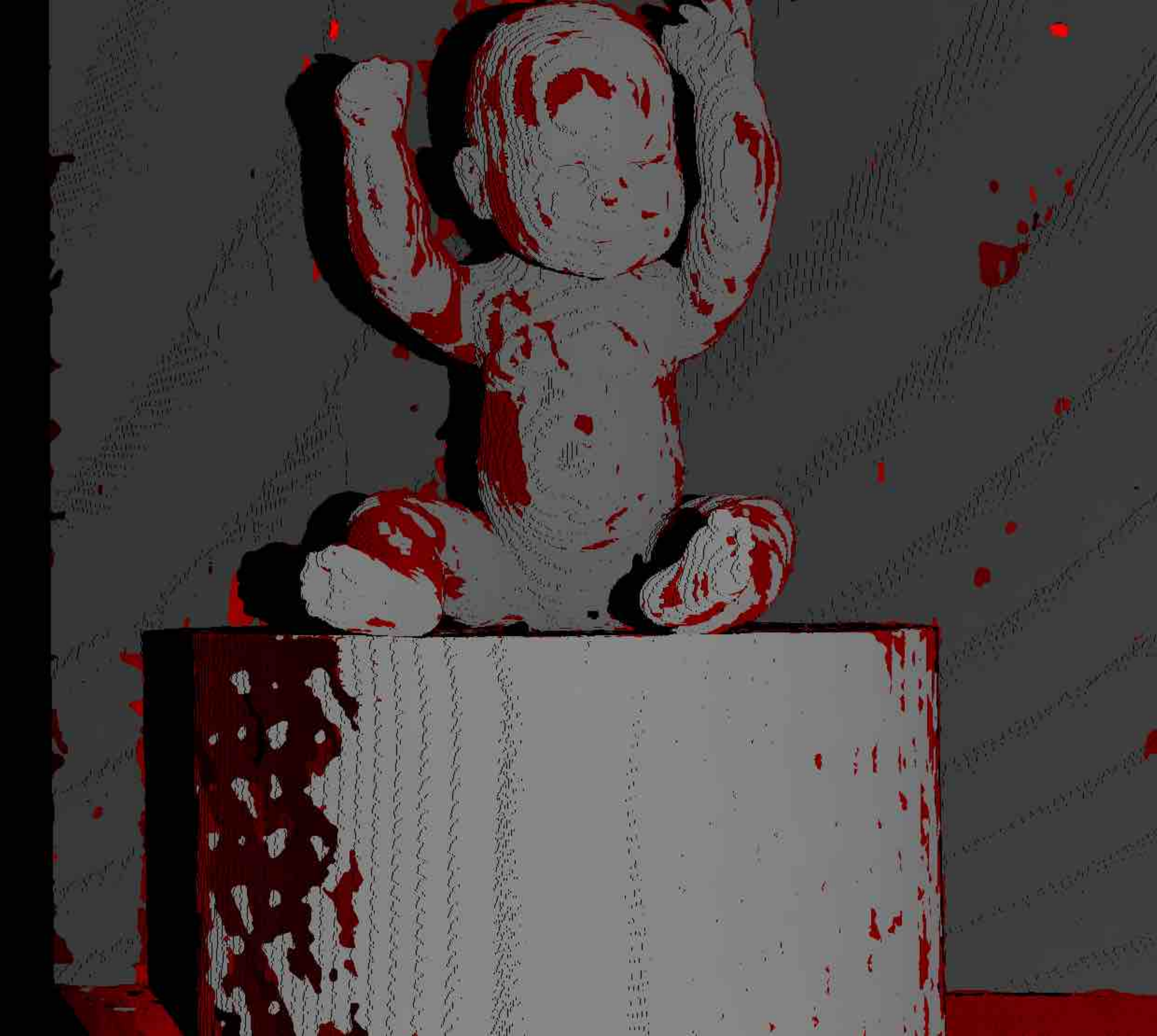}
	\end{minipage}
	\begin{minipage}[b]{0.155\linewidth}
		\center
		\includegraphics[width = 1.35cm]{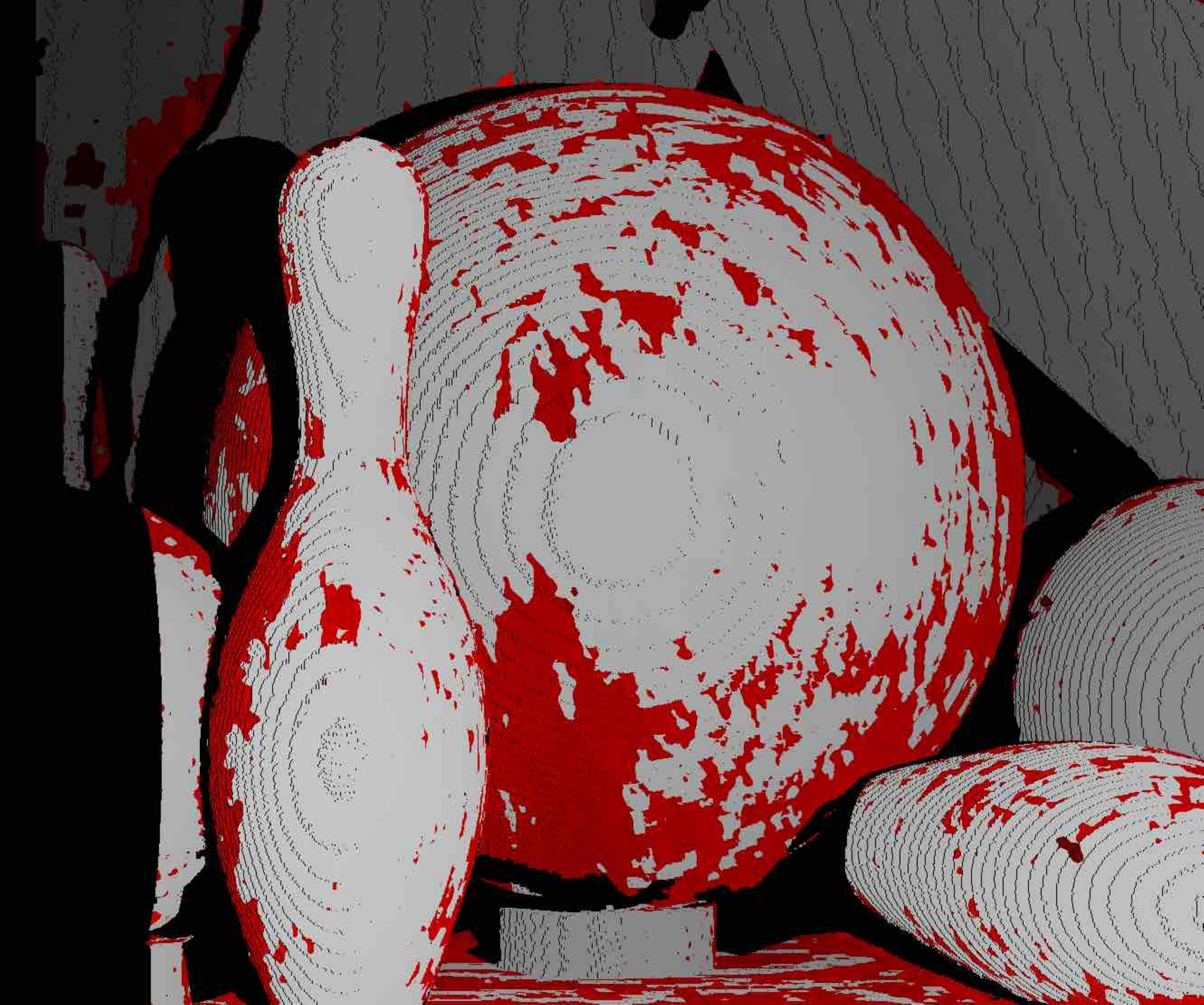}
	\end{minipage}
	\begin{minipage}[b]{0.17\linewidth}
		\center
		\includegraphics[width = 1.35cm]{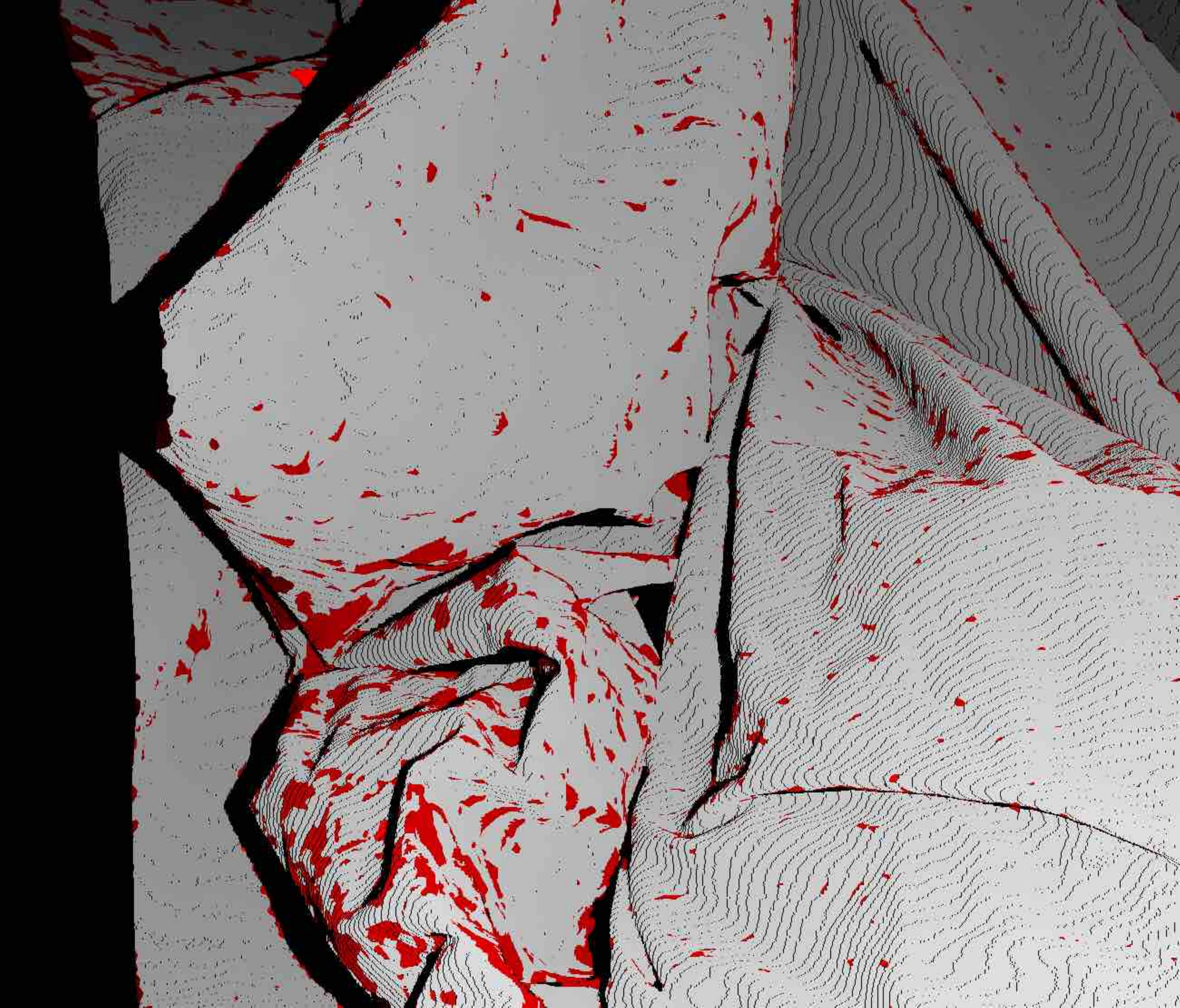}
	\end{minipage}
	\begin{minipage}[b]{0.155\linewidth}
		\center
		\includegraphics[width = 1.35cm]{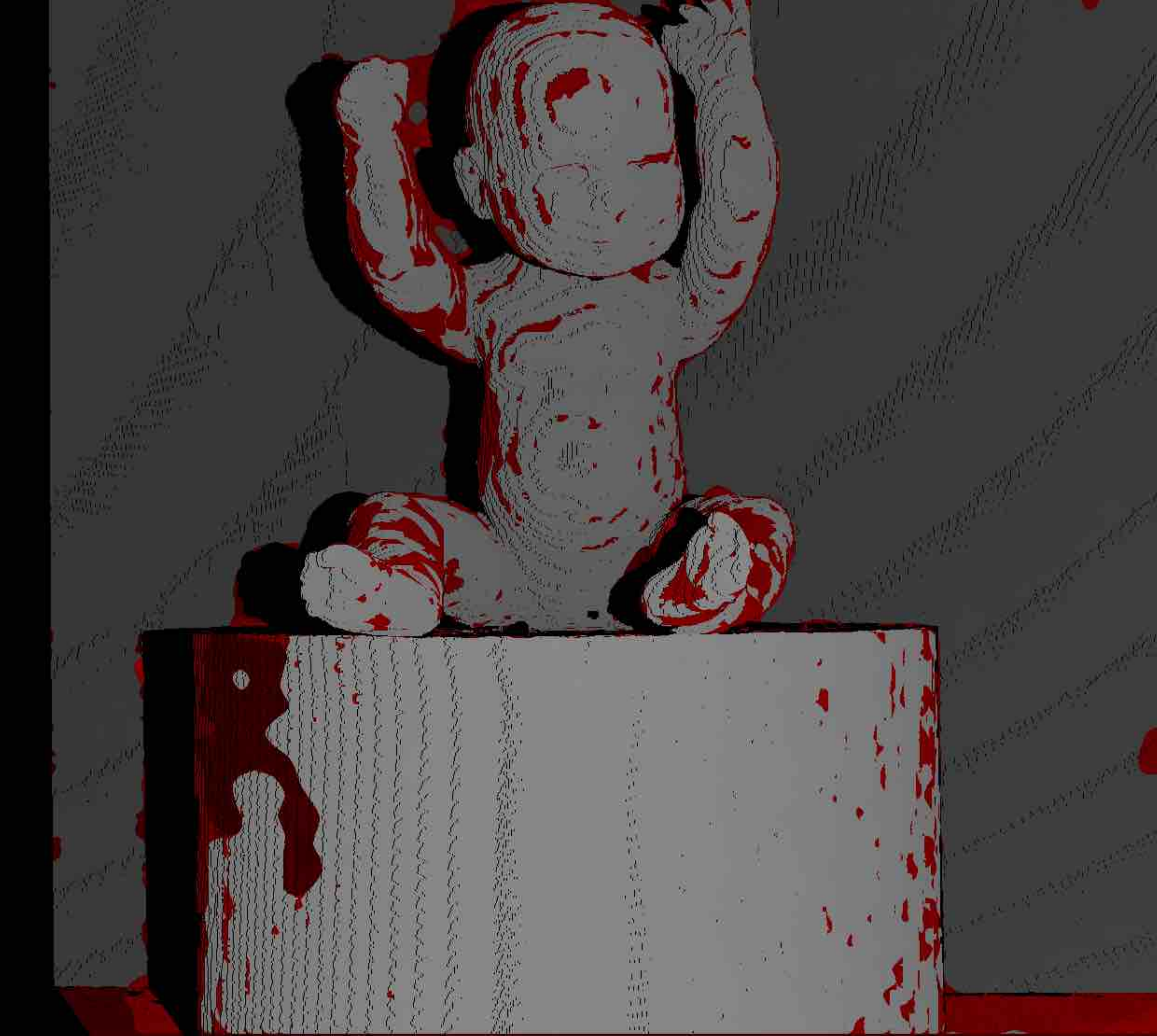}
	\end{minipage}
	\begin{minipage}[b]{0.155\linewidth}
		\center
		\includegraphics[width = 1.35cm]{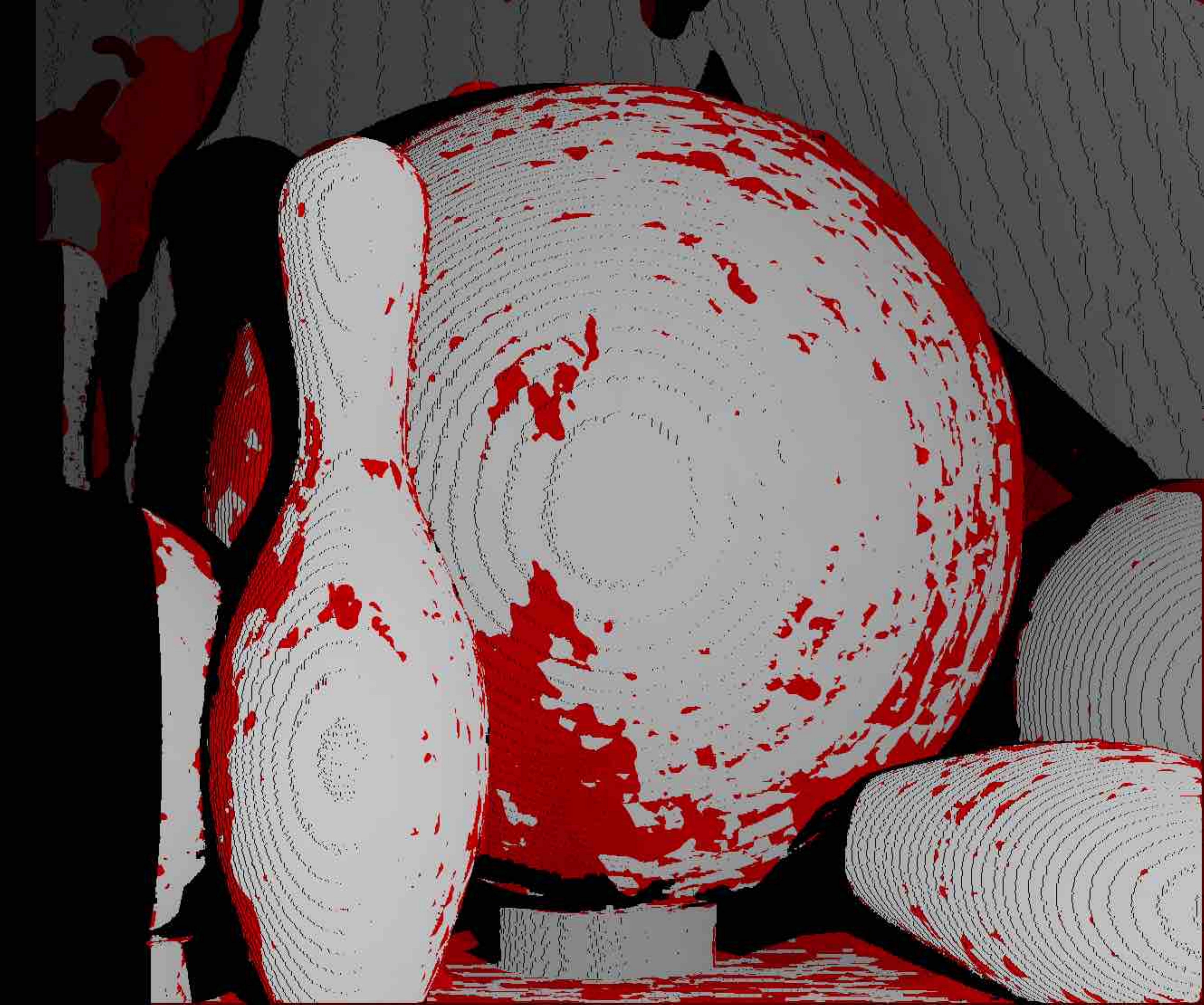}
	\end{minipage}
	\begin{minipage}[b]{0.155\linewidth}
		\center
		\includegraphics[width = 1.35cm]{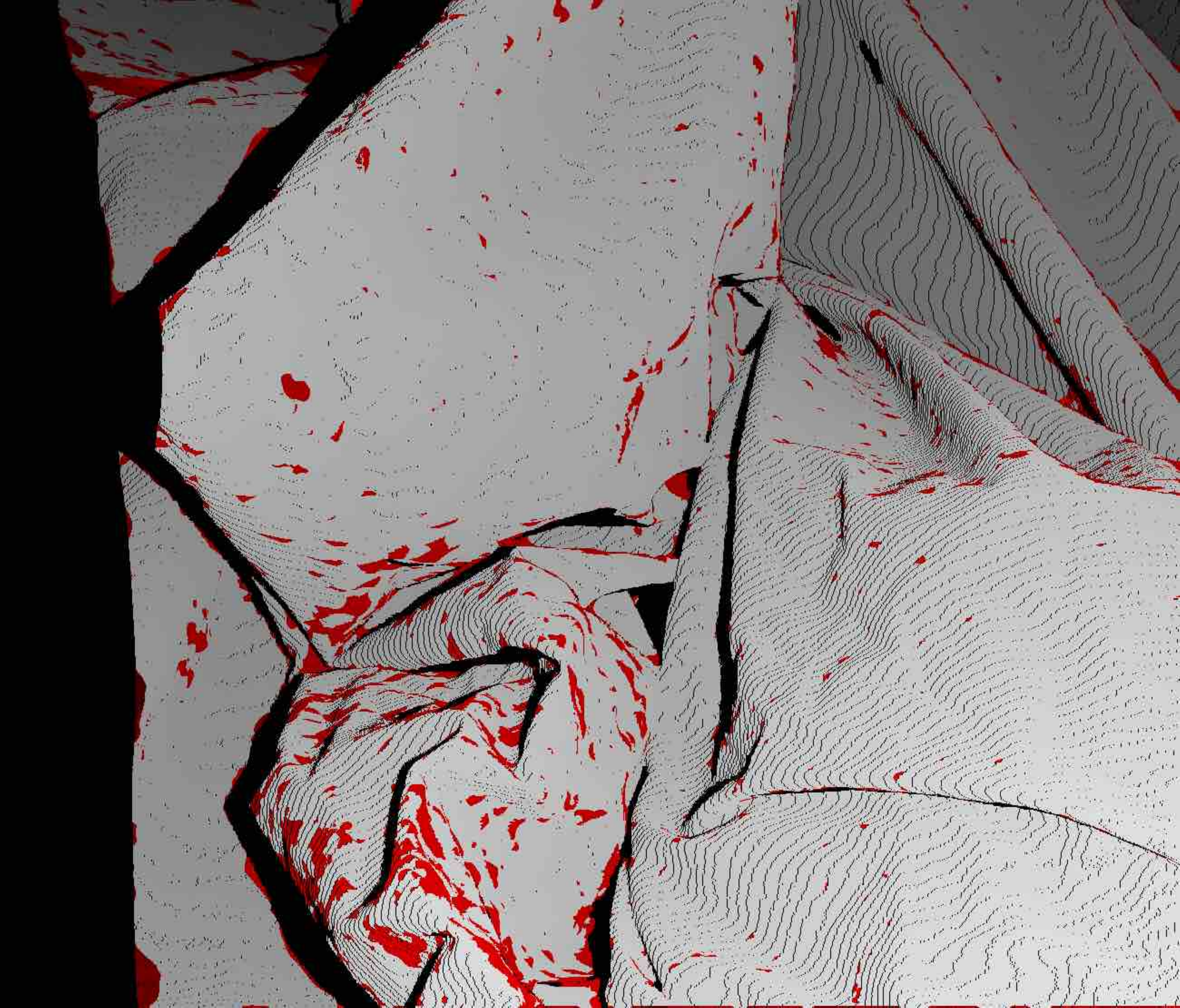}
	\end{minipage}

	\begin{minipage}[b]{0.155\linewidth}
		\center
		\includegraphics[width = 1.35cm]{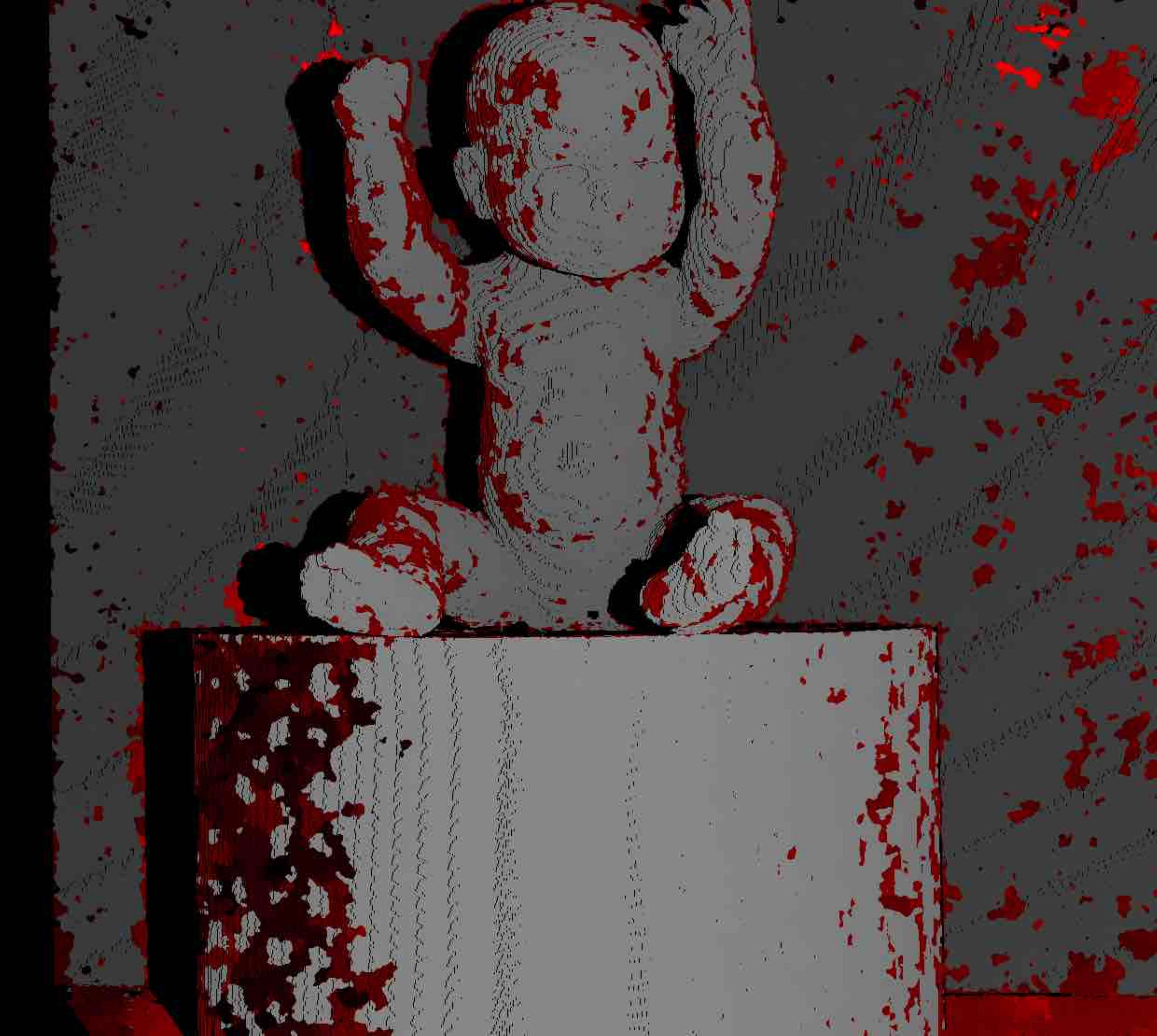}
	\end{minipage}
	\begin{minipage}[b]{0.155\linewidth}
		\center
		\includegraphics[width = 1.35cm]{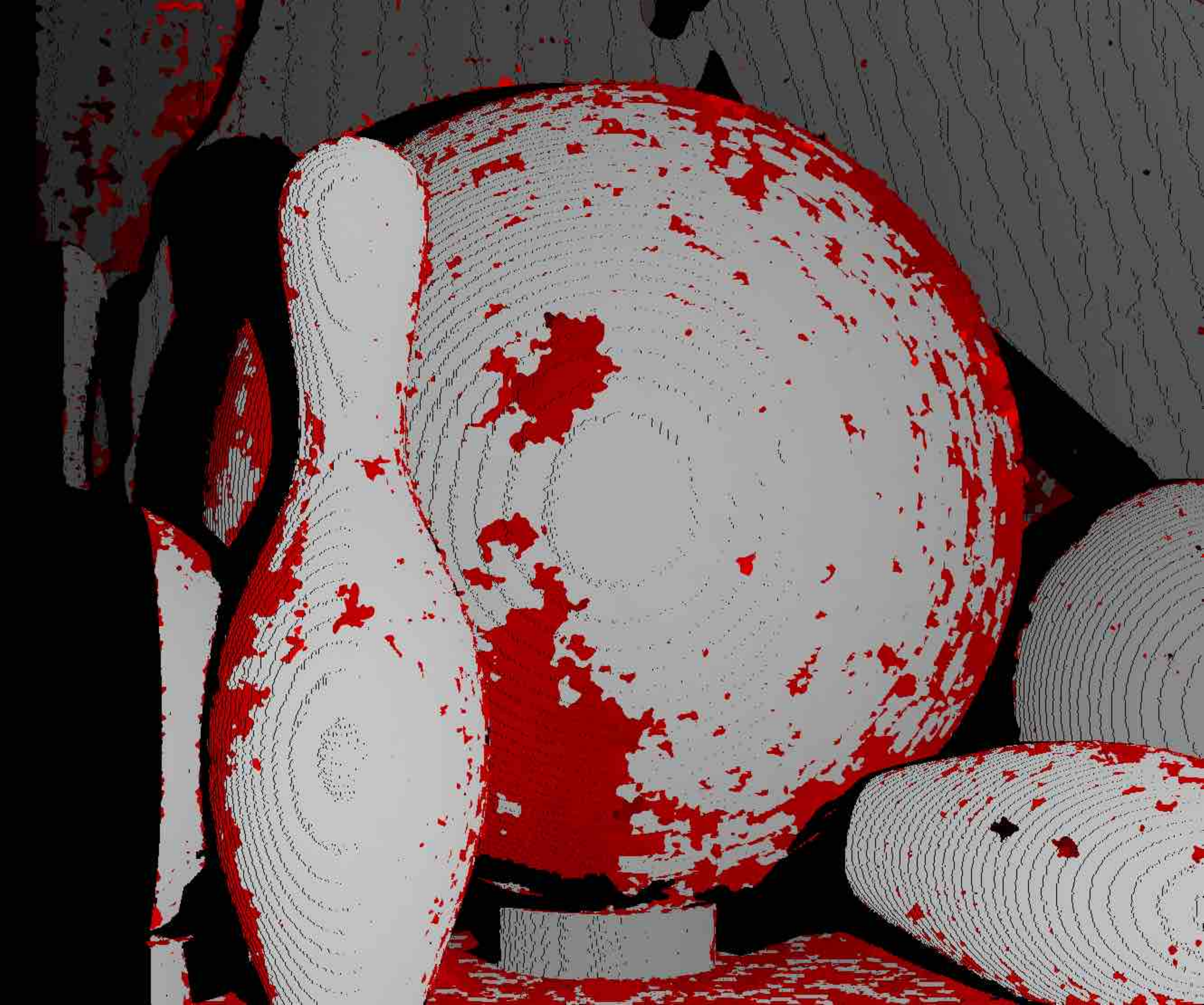}
	\end{minipage}
	\begin{minipage}[b]{0.17\linewidth}
		\center
		\includegraphics[width = 1.35cm]{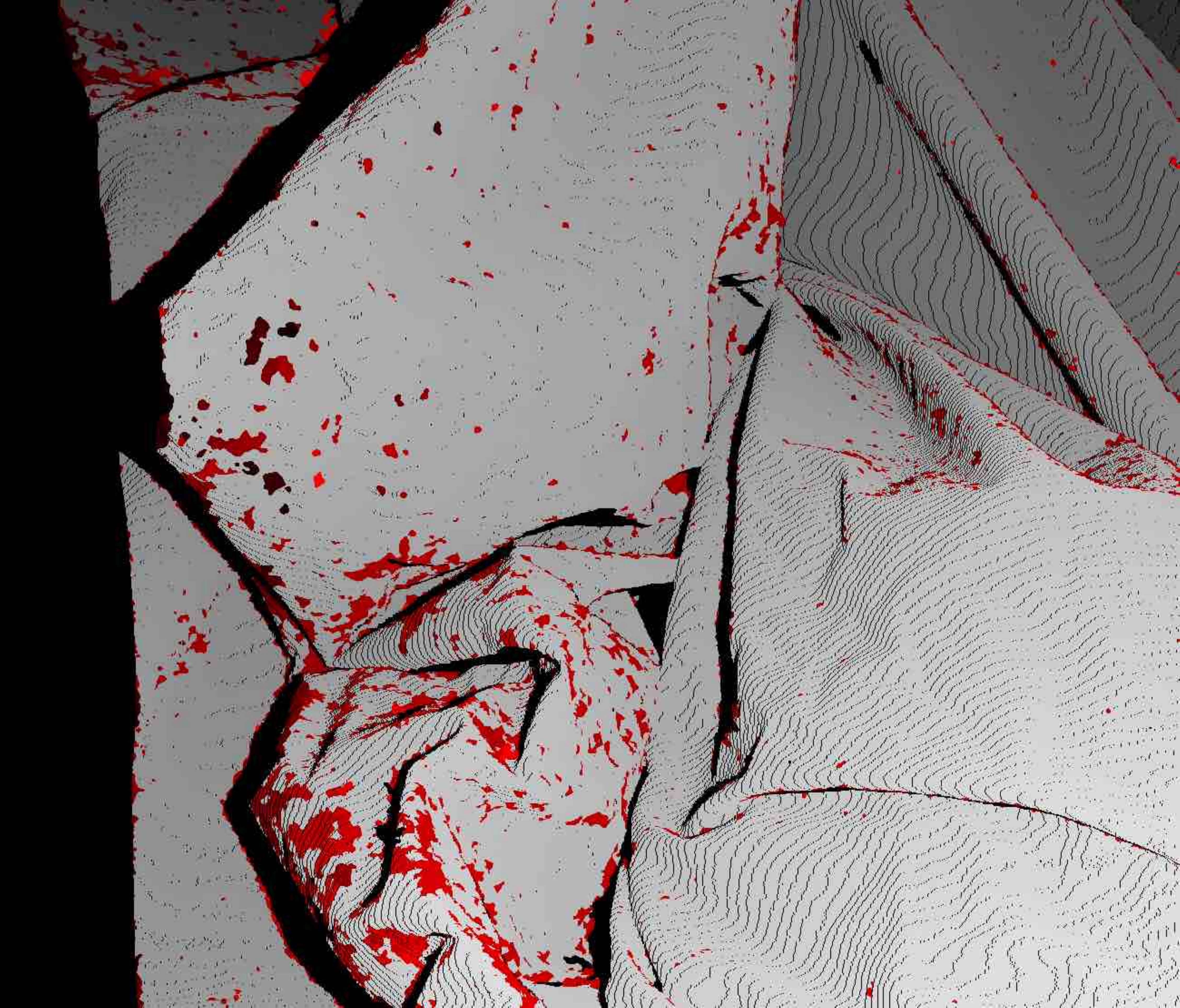}
	\end{minipage}
	\begin{minipage}[b]{0.155\linewidth}
		\center
		\includegraphics[width = 1.35cm]{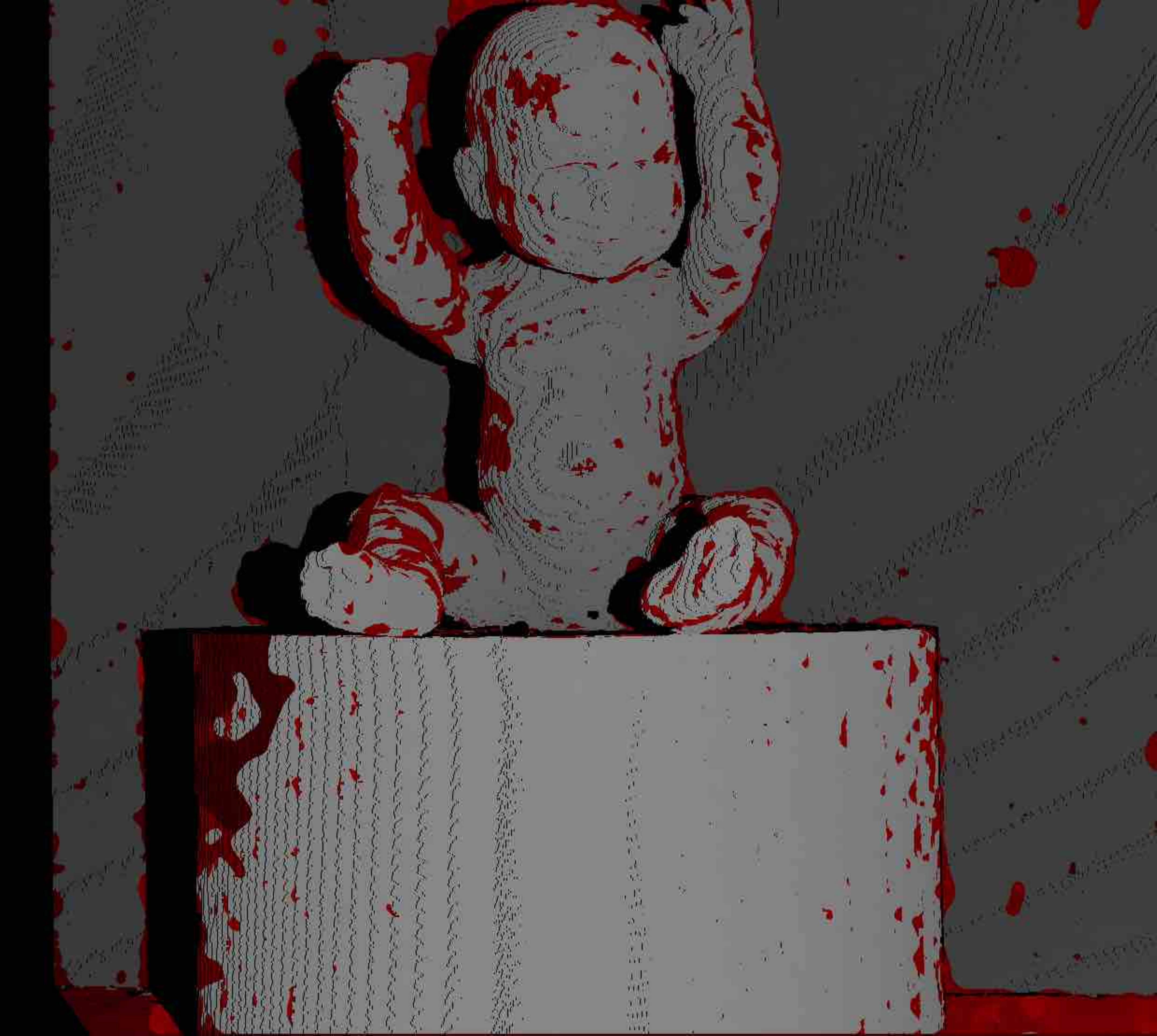}
	\end{minipage}
	\begin{minipage}[b]{0.155\linewidth}
		\center
		\includegraphics[width = 1.35cm]{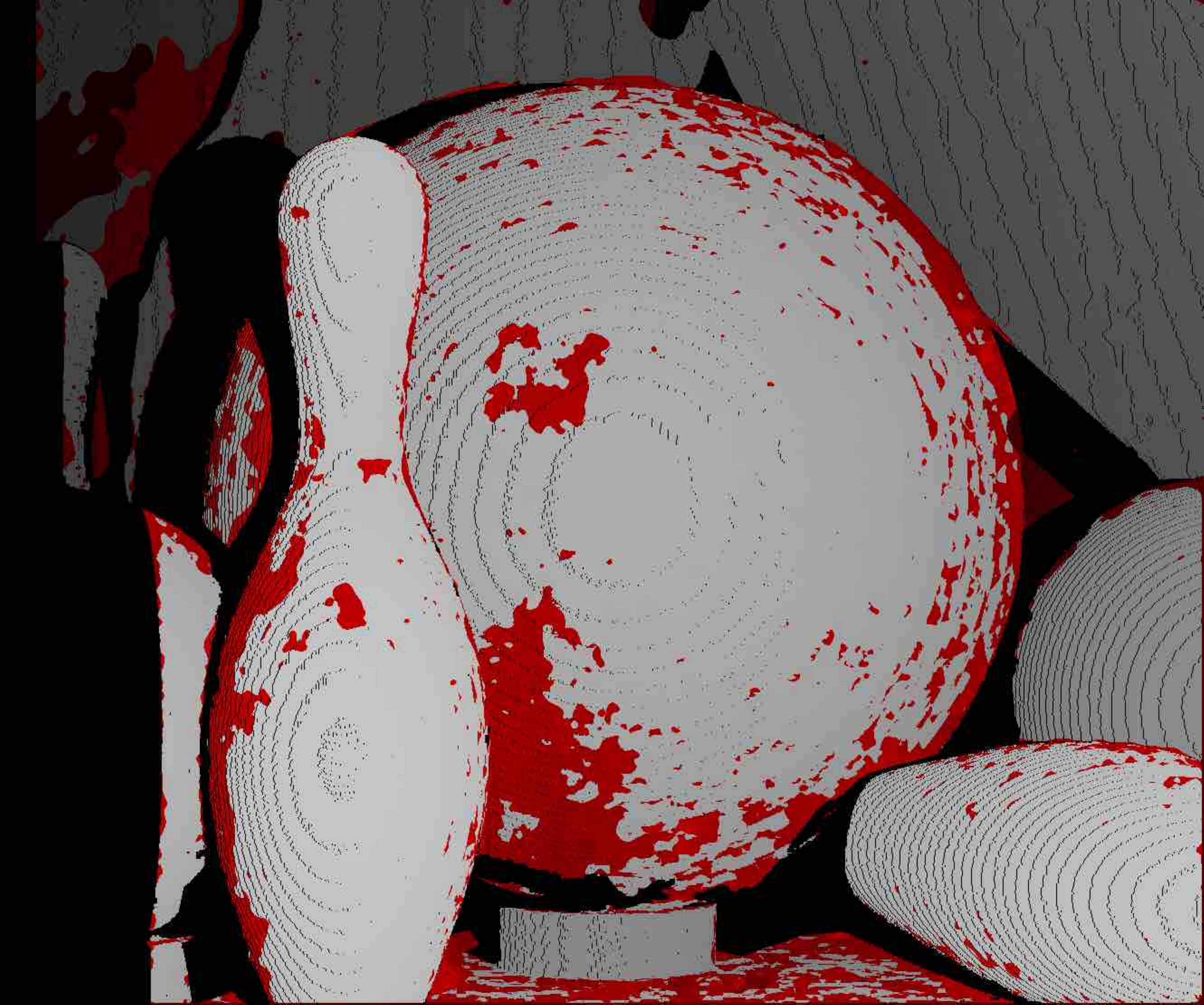}
	\end{minipage}
	\begin{minipage}[b]{0.155\linewidth}
		\center
		\includegraphics[width = 1.35cm]{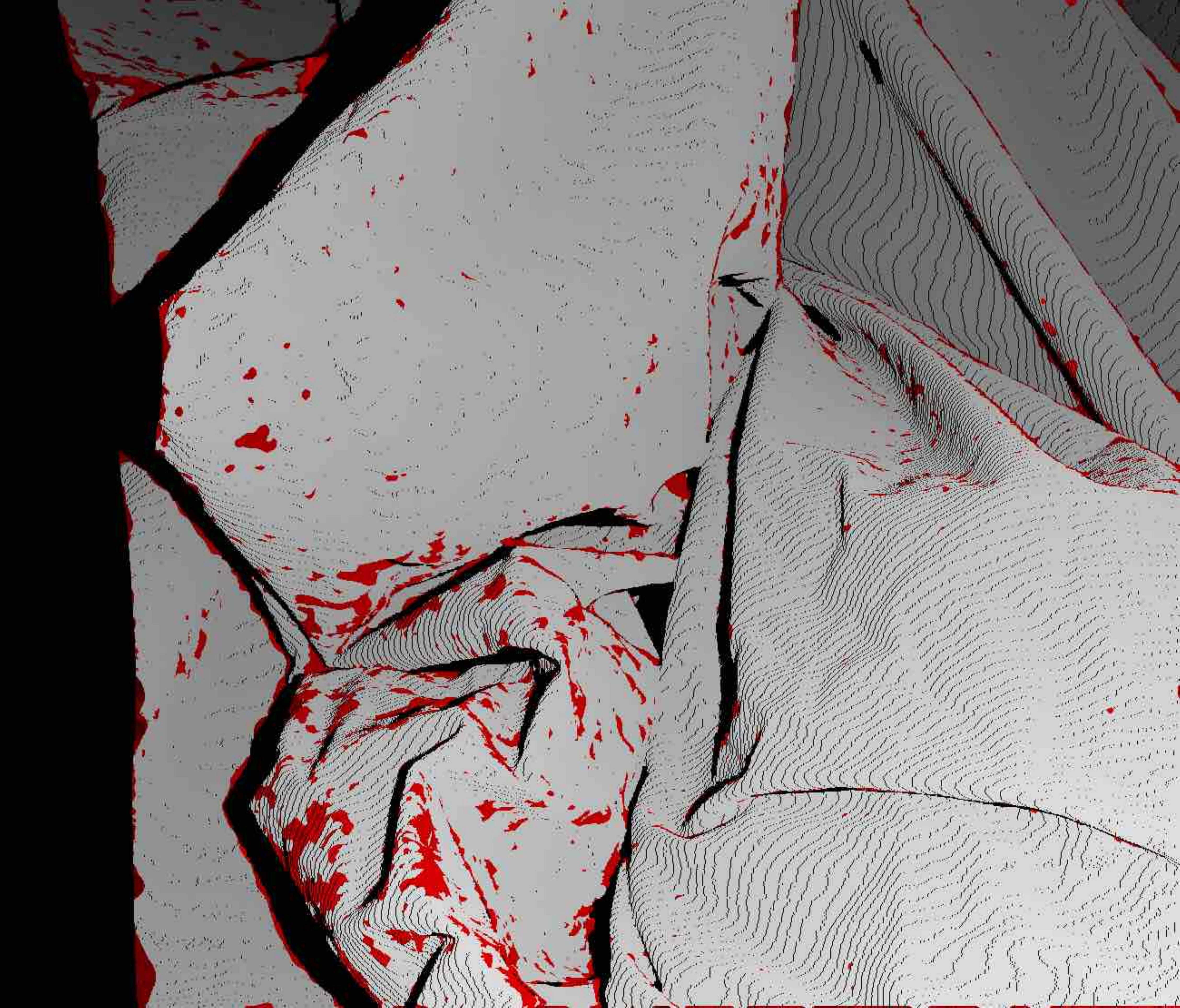}
	\end{minipage}
	\caption{Visual comparison of Baby1, Bowling2 and Cloth2. Left three: results from the original MST (first row), ST(second row) and RT (third row); Right three: results from the HDP+MST (first row), HDP+ST (second trow) and HDP+RT (third row).}
	\label{fig:middresult}
\end{figure}

 Fig.\ref{fig:middresult} presents the disparity maps obtained by different methods. Pixels with erroneous disparities are marked red and pixels in occluded regions are marked black. It can be seen that HDP-based algorithms achieve better accuracy in the less-textured regions where the tree-based algorithms fail. 
 This is because: \textbf{1)} evaluating the disparities in textureless regions on smaller images is easier and the more accurate disparity information is propagated down from the smallest layer to the largest one in our hierarchical graph pyramid structure;  \textbf{2)} the DTs well segment the graph according to disparity similarity directly; \textbf{3)} each DT is small enough to maintain detail information for preserving sharp depth discontinuities, and yet is large enough to contain as many pixels as possible with similar disparity intervals; \textbf{4)} pixels with very different disparities are in different DTs and thus have no impact on each other during aggregation.

 \paragraph{\bf Performance on KITTI}
The experiments in \cite{KITTI} show that methods with high ranking on established benchmarks such as Middlebury are ineffective in the real scenes due to the complex structured environments and large non-textured regions. In order to further evaluate the performance, we demonstrated our approach on KITTI benchmark for real-world images.
 
\begin{figure}[t]
 		\begin{minipage}[t]{0.49\linewidth}
 			\centerline{
                \includegraphics[width=\linewidth]{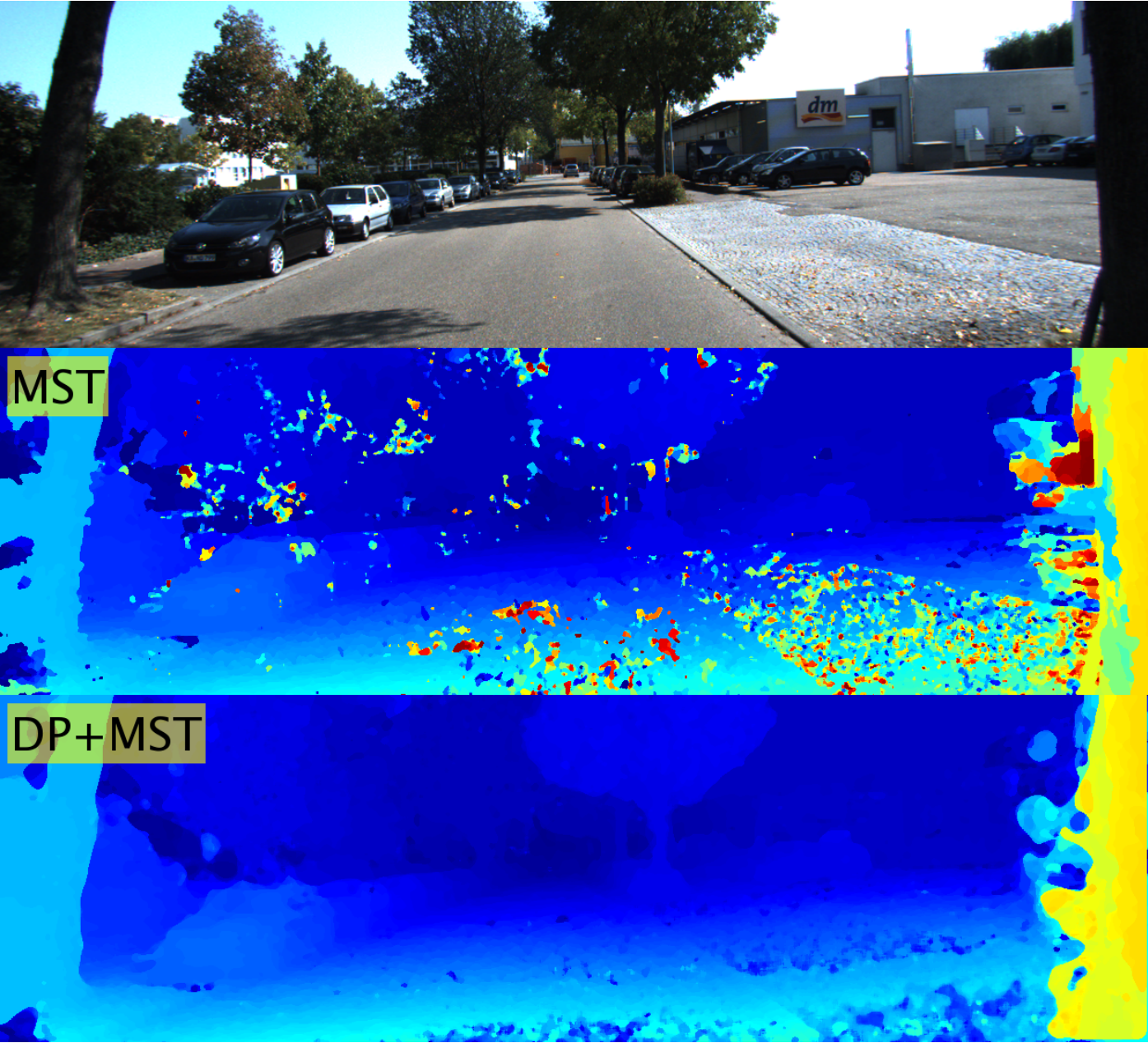}
 			}
 		\end{minipage}
 		\begin{minipage}[t]{0.49\linewidth}
 			\centerline{
        		\includegraphics[width=\linewidth]{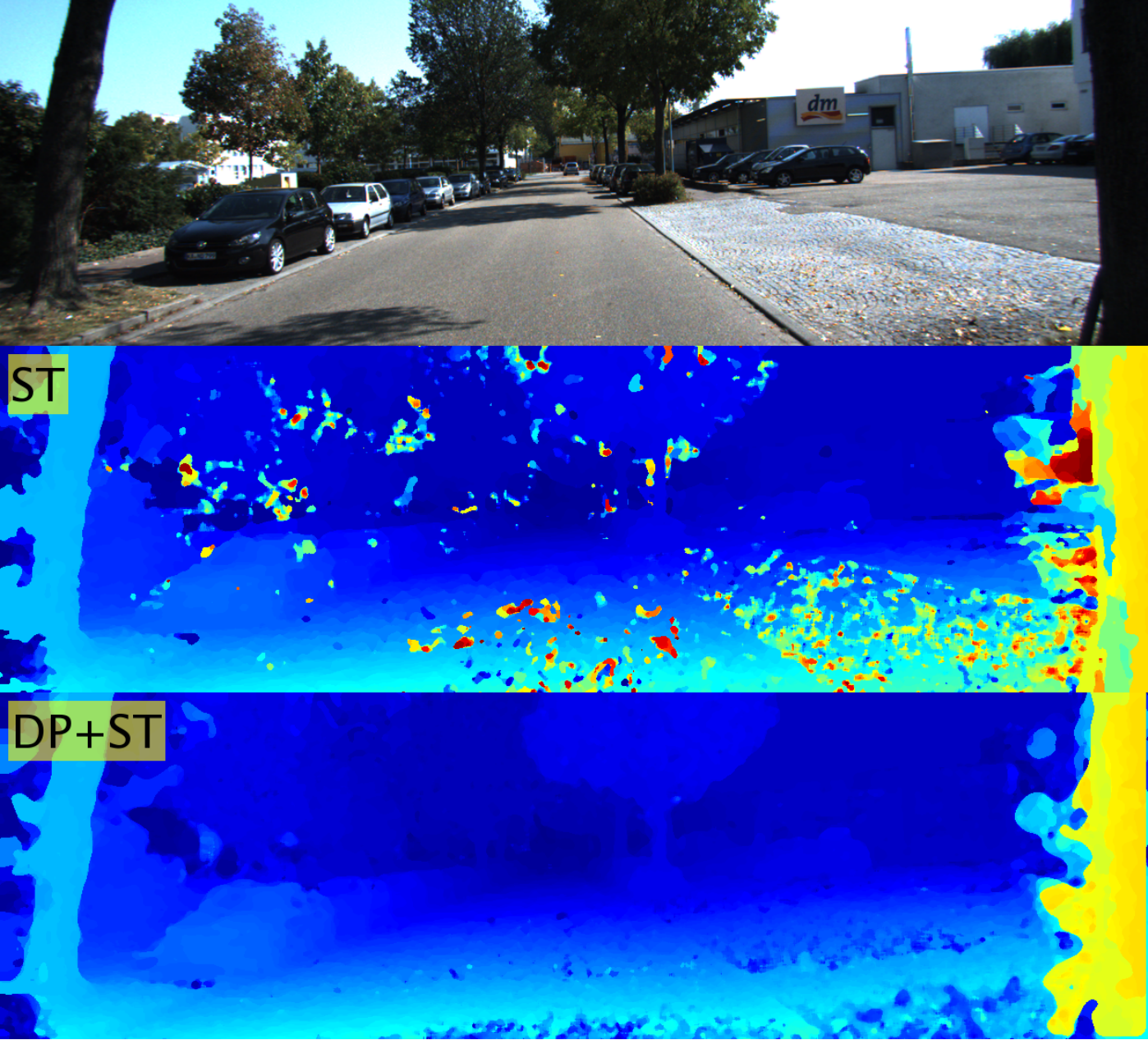}
 			}
 		\end{minipage}
 		\caption{Visual comparison between the MST, ST and their HDP versions on imageries in KITTI ($2011\_09\_26\_drive\_0009$).}
 		\label{fig:kittiresult}
 	\end{figure}

 Fig.~\ref{fig:kittiresult} presents the disparity maps for a typical high resolution imagery from KITTI. The results indicate less noise in the disparity maps calculated by our algorithm but the original tree-based methods fail due to the lack of texture, especially in large non-texture regions (e.g. sky and roads). The complete results over the KITTI ``$2011\_09\_26\_drive\_0009$" sequence can be found in the supplementary materials.
  	
  	\paragraph{\bf Performance on Low-quality dataset}
 	
\begin{figure}
 				\includegraphics[width=\linewidth]{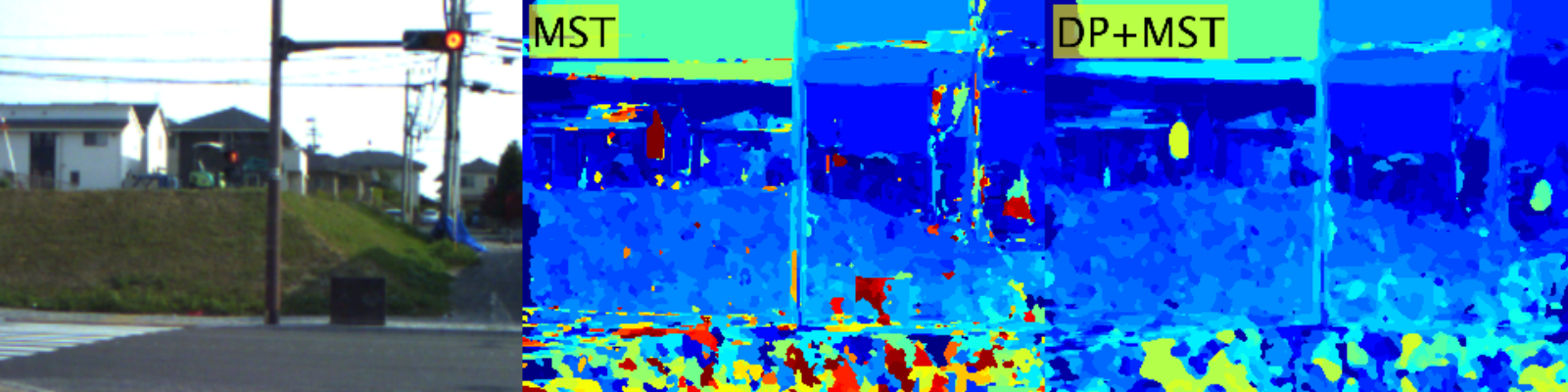}
 			\center
 				\includegraphics[width = \linewidth]{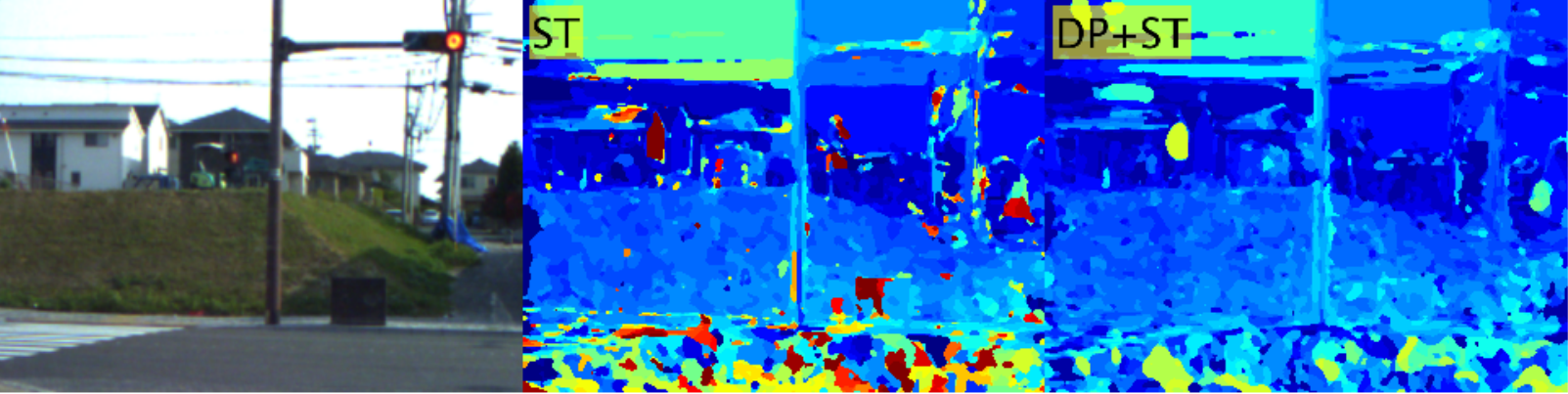}
 		\caption{Visual comparison between the MST, ST and their HDP versions on a typical image in Low-quality dataset.}
 		\label{fig:wasedaresult}
 	\end{figure}

 	It is also crucial to evaluate the proposed method with low quality images. Fig.\ref{fig:wasedaresult} presents the disparity maps computed by the MST, ST and their HDP versions for a typical low quality images. The proposed aggregation strategy reduces the influence of low image quality since more accurate disparity maps are computed. Compared with MST and ST, there is less noise in the disparity maps generated by our method on the large non-textured regions.
 	 	
 \section{Conclusion}
 \label{sec:conclu}
 This paper aims to reduce the computation cost of tree-based stereo matching algorithms caused by the large disparity search range. We propose a hierarchical disparity prediction model to significantly reduce the disparity interval, which predicts the disparity of the lower layer graph from that of the upper layer in a graph pyramid constructed from the original image. Some independent disparity trees are constructed from pixels with similar disparity vlaues in each layer. The cost aggregartion is conducted on the disparity trees. The proposed hierarchical disparity prediction model can be easily incoporated into the existing tree-based stereo mathcing frameworks and compared to the original algorithms, the ones combining our hierarchical disparity prediction are not only more efficient but also improve depth estimation results. Moreover, this framework also removes the negative effect of image's low quality on the result to some extent. This amazing performance is achieved by predicting credible disparity intervals, which not only allow the computation to be carried out within small intervals, but also well segment the graph based on the disparities. Smaller disparity intervals give rise to the speedup, and the depth-based segmentation leads to the higher accuracy.
 
 To sum up, our contributions include three parts:
 \begin{enumerate}
 	\item We propose a hierarchical disparity prediction model to predict the distribution of disparities in the lower layer of a graph pyramid from its upper layer.
 	\item We propose an efficient method to generate the small disparity intervals.
 	\item We propose a general acceleration framework for the existing tree-based algorithms to make them faster and more accurate.
 \end{enumerate}

{\small
\bibliographystyle{ieee}
\bibliography{mybibfile}
}

\end{document}